\documentclass[journal]{IEEEtran}
\usepackage{amsmath,amsfonts}
\usepackage{algorithmic}
\usepackage{algorithm}
\usepackage{array}
\usepackage{textcomp}
\usepackage{stfloats}
\usepackage{url}
\usepackage{verbatim}
\usepackage{graphicx}
\usepackage{subcaption}
\usepackage{overpic}
\usepackage{cite}
\usepackage{multirow}
\usepackage{amsmath}
\usepackage{makecell}
\usepackage{lscape}
\usepackage[justification=centering]{caption}
\usepackage[colorlinks,linkcolor=blue,anchorcolor=blue,citecolor=blue]{hyperref}
\usepackage{colortbl}
\usepackage[table,xcdraw]{xcolor}
\usepackage{float}
\usepackage{epstopdf}
\usepackage{textcomp}
\usepackage{bbding}
\usepackage{booktabs}
\usepackage[normalem]{ulem}
\useunder{\uline}{\ul}{}
\usepackage{indentfirst}
\definecolor{Gray}{gray}{0.9} % 定义浅灰色

\begin{document}
\title{SGHA-Attack: Semantic-Guided Hierarchical Alignment for Transferable Targeted Attacks on Vision-Language Models}

\author{Haobo~Wang,
        Weiqi Luo,  ~\IEEEmembership{Senior Member,~IEEE,}
        Xiaojun Jia, 
        Xiaochun Cao, ~\IEEEmembership{Senior Member,~IEEE}

%\thanks{This work was supported by the National Natural Science Foundation of China (Grant No. 62472458), the Natural Science Foundation of Guangdong Province (Grant No. 2025A1515012823) and the Project of Guangdong Provincial Key Laboratory of Information Security Technology (Grant No.2023B1212060026). Corresponding author: Weiqi Luo.}
\thanks{Haobo Wang and Weiqi Luo are with Guangdong Province Key Lab of Information Security Technology,  and School of Computer Science and Engineering, Sun Yat-sen University, Guangdong 510006, China (e-mail:wanghb69@mail2.sysu.edu.cn; luoweiqi@mail.sysu.edu.cn.) Corresponding author: Weiqi Luo.} 
\thanks{Xiaojun Jia is with Nanyang Technological University, Singapore (e-mail: jiaxiaojunqaq@gmail.com).}
\thanks{Xiaochun Cao is with the School of Cyber Science and Technology, Shenzhen Campus, Sun Yat-sen University, Shenzhen 518107, China. (e-mail:caoxiaochun@mail.sysu.edu.cn)}

%%(e-mail:wanghb69@mail2.sysu.edu.cn; luoweiqi@mail.sysu.edu.cn; xiexiaoh6@mail.sysu.edu.cn; zhpj@mail.sysu.edu.cn; huangwm36@mail2.sysu.edu.cn.)} 
%\thanks{Jiwu Huang is with the Guangdong Laboratory of Machine Perception and  Intelligent Computing, Faculty of Engineering, Shenzhen MSU-BIT University, Shenzhen, 518116, China (E-mail: jwhuang@smbu.edu.cn).}
}

%% The paper headers
%\markboth{Journal of \LaTeX\ Class Files,~Vol.~14, No.~8, August~2021}%
%{Shell \MakeLowercase{\textit{et al.}}: A Sample Article Using IEEEtran.cls for IEEE Journals}

\maketitle 

\begin{abstract}
Large vision–language models (VLMs) are vulnerable to transfer-based adversarial perturbations, enabling attackers to optimize on surrogate models and manipulate black-box VLM outputs. Prior targeted transfer attacks often overfit surrogate-specific embedding space by relying on a single reference and emphasizing final-layer alignment, which underutilizes intermediate semantics and degrades transfer across heterogeneous VLMs. To address this, we propose SGHA-Attack, a Semantic-Guided Hierarchical Alignment framework that adopts multiple target references and enforces intermediate-layer consistency. Concretely, we generate a visually grounded reference pool by sampling a frozen text-to-image model conditioned on the target prompt, and then carefully select the Top-K most semantically relevant anchors under the surrogate to form a weighted mixture for stable optimization guidance. Building on these anchors, SGHA-Attack injects target semantics throughout the feature hierarchy by aligning intermediate visual representations at both global and spatial granularities across multiple depths, and by synchronizing intermediate visual and textual features in a shared latent subspace to provide early cross-modal supervision before the final projection. Extensive experiments on open-source and commercial black-box VLMs show that SGHA-Attack achieves stronger targeted transferability than prior methods and remains robust under preprocessing and purification defenses.
\end{abstract}


\section{Introduction}

\begin{figure}[t]
  \centering
  \includegraphics[width=\linewidth]{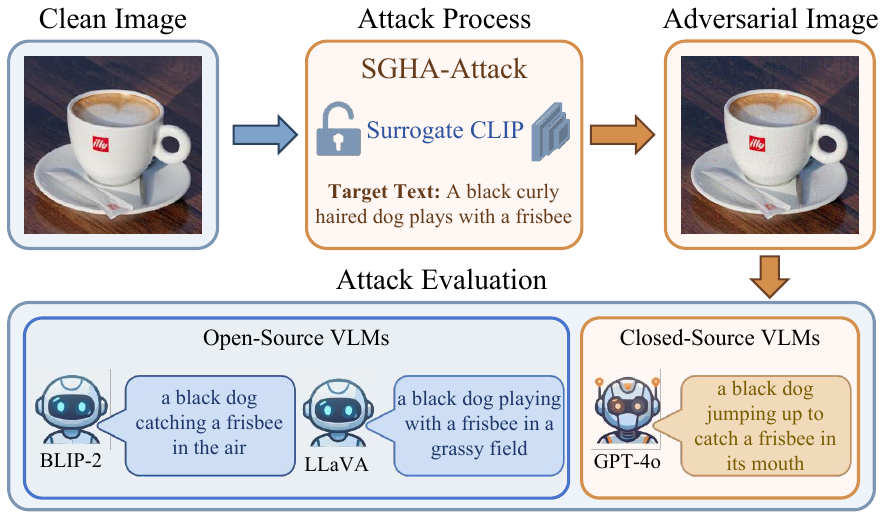}
  \caption{Illustration of the proposed SGHA-Attack framework targeting diverse VLMs.}
  \label{fig:attack_demo}
\end{figure}

\IEEEPARstart{L}{arge} Vision-Language Models ~\cite{VLMsurvey} have emerged as a foundational technology, powering critical applications from autonomous agents to content moderation~\cite{BLIP,LLaVA,instruction_following}. As these systems are increasingly entrusted with high-stakes decision-making~\cite{DriveGPT4,moor2023foundation}, their inherent vulnerability to adversarial perturbations poses severe security risks~\cite{VLMsurvey2}. Beyond causing benign classification errors~\cite{BardAttack}, imperceptible perturbations can effectively hijack a model's semantic understanding, enabling adversaries to circumvent safety guardrails (i.e., jailbreaking) and elicit policy-violating outputs~\cite{VisualJailbreak}. In the black-box scenario, targeted transfer-based attacks~\cite{TransferabilitySurvey} represent a particularly practical threat. By optimizing on an accessible surrogate, attackers can precisely steer proprietary commercial APIs to generate specific target captions without knowledge of the victim's internal parameters. This transferability threatens the integrity of downstream ecosystems, including automated decision support~\cite{IndirectPromptInjection} and tool-augmented agentic systems~\cite{MultimodalAgentAttacks}. To unveil the severity of this threat, we introduce {SGHA-Attack}, as illustrated in Fig.~\ref{fig:attack_demo}, a novel framework designed to effectively mislead diverse black-box VLMs to generate attacker-specified target captions.

Existing transfer-based attacks can be broadly categorized into perturbation-constrained methods and unrestricted diffusion-based approaches. In the constrained category, methods like AttackVLM~\cite{AttackVLM} optimize adversarial perturbations by aligning global image embeddings with a target text embedding within a restricted $\ell_\infty$ budget. Chain-of-Attack (COA)~\cite{COA} further enhances this by incorporating modality fusion and iterative optimization strategies to improve semantic consistency. M-Attack~\cite{M-Attack} introduces robust input transformations, such as random resizing and cropping, to bypass the preprocessing pipelines of commercial models. In the unrestricted category, approaches like AdvDiffVLM~\cite{AdvDiffVLM} leverage latent diffusion models to generate adversarial examples. Specifically, they utilize attention-guided masking to preserve critical features, fundamentally operating by reconstructing image content rather than adding imperceptible noise. Despite different designs, these transfer-based attacks have two common limitations. First, they rely on a single target reference, which can be sensitive to the surrogate model and thus transfer poorly to other VLMs. Second, they match embedding similarity only at the surrogate encoder’s final layer, ignoring intermediate features, this can cause overfitting to the surrogate and weaken targeted transfer to instruction-tuned or proprietary VLMs.

%缺少了一个缺陷，全过程如何解决对应一下limitation，Forthemore. （feature extraction process） 
To address these limitations, we propose SGHA-Attack, a \emph{Semantic-Guided Hierarchical Alignment} framework that enforces target consistency across intermediate-layer representations rather than only at the final output.
To tackle the single-target limitation, we introduce Semantic-Guided Anchor Injection (SGAI). Instead of optimizing toward one target reference, SGAI uses multiple semantic anchors that capture diverse realizations of the target concept and forms a weighted mixture of these anchors as the optimization target, making the guidance more stable and improving targeted transfer across different VLMs.
To address the second limitation of relying on final-layer matching, we further perform hierarchical alignment over intermediate representations: Hierarchical Visual Structure Alignment (HVSA) constrains layer-wise visual features at multiple intermediate depths, aligning both global and spatial structure, while Cross-Modal Latent Space Synchronization (CLSS) provides intermediate cross-modal supervision in a shared latent subspace by encouraging projected visual features to remain aligned with the evolving textual semantics rather than only at the final output. Together, these designs provide multi-layer supervision and yield adversarial examples with stronger targeted transferability across a wide range of open-source and commercial black-box VLMs.

%Our main contributions are summarized as follows:
%\begin{itemize}
%    \item We analyze limitations of late-alignment objectives in transfer-based targeted attacks on VLMs, and propose SGHA-Attack to enforce semantic consistency via hierarchical, multi-granularity constraints.
%    \item We introduce {SGAI}, {HVSA}, and {CLSS} to provide {diverse and semantically complementary} visually grounded priors and exploit intermediate representations of the surrogate encoder, offering cross-modal supervision beyond final-layer embedding matching.
%    \item Extensive experiments on both open-source and commercial black-box VLMs show that SGHA-Attack outperforms prior methods in targeted transferability and demonstrates improved resilience under preprocessing and purification defenses.
%\end{itemize}

Our main contributions are summarized as follows:
\begin{itemize}
    \item We propose SGHA-Attack, a Semantic-Guided Hierarchical Alignment framework for transfer-based targeted attacks on VLMs that moves beyond late-stage embedding matching by enforcing hierarchical, multi-granularity semantic consistency.
    \item We introduce SGAI to build {multiple} target references through {generation and careful selection} (generate an anchor pool and select Top-$K$), and leverage HVSA/CLSS to align intermediate visual features and synchronize intermediate cross-modal tokens.
    \item Extensive experiments on open-source and commercial black-box VLMs show that SGHA-Attack improves targeted transferability over prior methods and remains robust under preprocessing and purification defenses.
\end{itemize}

\section{Related Work}
\label{sec:related}

\subsection{Vision--Language Models}

VLMs have evolved from task-specific architectures to more general-purpose, generative foundation models~\cite{GPT,Gemini,VLMsurvey}. Prior to the widespread adoption of large language models (LLMs), many VLMs emphasized learning joint visual--textual representations for discriminative objectives, including image--text retrieval~\cite{I2T-Retrieval-1,I2T-Retrieval-2} and visual question answering~\cite{VQA1,VQA2}. More recently, the emergence of strong LLMs has shifted the dominant paradigm toward leveraging an LLM as a language and reasoning backbone while introducing a visual front-end that enables the model to condition on images~\cite{LLMsurvey1,LLMsurvey2}. Under this paradigm, a central technical challenge is {modality alignment}: mapping visual representations into a form that an LLM can effectively consume and integrate with text~\cite{LLaMA,Vicuna}.

In practice, modality alignment is commonly implemented via a learnable projector (or adapter) that connects the visual encoder to the LLM. Two representative design families are widely adopted. The first family, exemplified by LLaVA~\cite{LLaVA} and MiniGPT-4~\cite{Minigpt-4}, uses lightweight mappings such as linear layers or MLPs to project visual tokens into the LLM embedding space, offering simplicity and computational efficiency. The second family, used by BLIP-2~\cite{BLIP2} and InstructBLIP~\cite{InstructBLIP}, introduces a query-based module (e.g., Q-Former) that employs learnable queries and cross-attention to selectively aggregate visual information before interfacing with the LLM. In this work, we study adversarial transferability across these architectures, with particular attention to how their alignment modules and intermediate representations influence vulnerability and generalization of attacks.

\subsection{Adversarial Attacks on VLMs}

Adversarial attacks aim to mislead models via imperceptible input perturbations. Depending on the adversary's knowledge, they are classified as {white-box} (full parameter access)~\cite{FGSM,CW} or {black-box} (no access to model parameters or gradients)~\cite{OPS,Admix}. Regarding objectives, attacks distinguish between {untargeted} ones that merely degrade performance and {targeted} ones that steer outputs toward specific malicious goals, posing greater safety risks. In realistic black-box scenarios, {transfer-based} attacks~\cite{Di,Adv-Inversion,FOA} have emerged as the dominant paradigm over query-based methods~\cite{PatchAttack}, as the latter are often impractical for large VLMs due to prohibitive inference latency and API costs. Consequently, our work focuses on transfer-based targeted attacks, which optimize perturbations on an accessible surrogate (e.g., CLIP) to compromise unseen victim VLMs via cross-model transferability.

Early research on VLM robustness primarily focused on untargeted degradation~\cite{untarget1,untarget2,untarget3,untarget4,untarget5}. Recently, significant progress has been made in targeted transfer-based attacks. Representative works like AttackVLM~\cite{AttackVLM} craft targeted examples by aligning global image embeddings with a specific target reference. Subsequent methods such as Chain-of-Attack (COA)~\cite{COA} improve semantic consistency through iterative optimization and modality fusion, while IPGA~\cite{IPGA} exploits intermediate projector representations (e.g., Q-Former) for fine-grained alignment.  Others, including AdvDiffVLM~\cite{AdvDiffVLM} and M-Attack~\cite{M-Attack}, further enhance transferability via diffusion-based generation or robust input transformations. However, these methods predominantly operate on the encoder's output or the subsequent projection stage, treating the {visual backbone itself} as a monolithic feature extractor. This strategy overlooks the rich hierarchical representations within the visual encoder, where features evolve from low-level structures to high-level semantics. In contrast, our work explicitly opens the visual backbone and {synchronizes its intermediate hierarchical features with corresponding textual representations}, thereby enforcing deep semantic consistency throughout the entire feature extraction process.

\begin{figure*}[t]
  \centering
  \includegraphics[width=\textwidth]{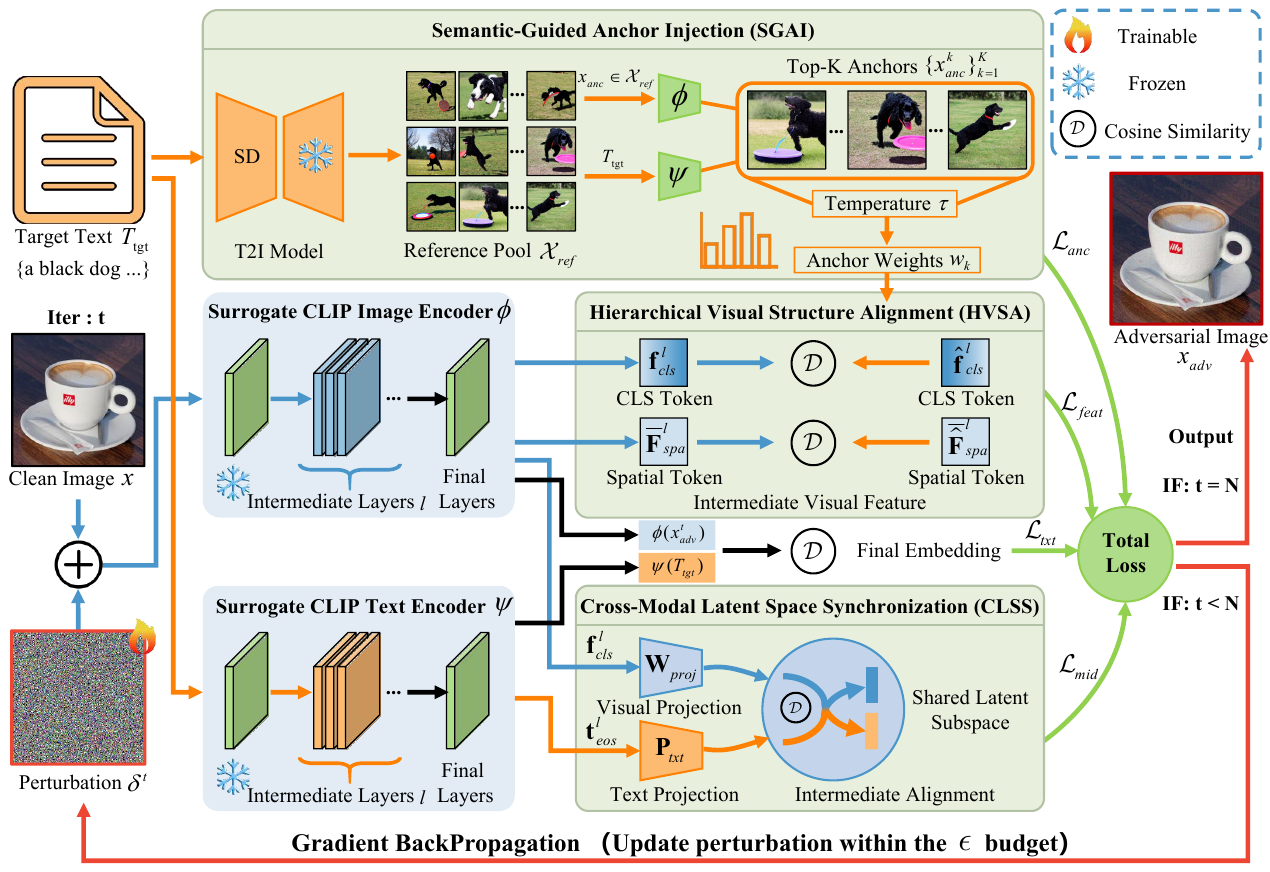}
  \caption{Overview of the proposed SGHA-Attack framework for generating transferable targeted adversarial examples. The framework consists of three key components: SGAI, HVSA, and CLSS.}
\label{fig:attack_total}
\end{figure*}

%\section{Methodology}
%\label{sec:method}
%
%As illustrated in Fig.~\ref{fig:attack_total}, we present SGHA-Attack, a transfer-based targeted attack framework designed to craft an adversarial example $\boldsymbol{x}_{\mathrm{adv}}$ by adding a human-imperceptible perturbation $\boldsymbol{\delta}$ to a clean image $\boldsymbol{x}$, formulated as $\boldsymbol{x}_{\mathrm{adv}} = \boldsymbol{x} + \boldsymbol{\delta}$ under the constraint $\|\boldsymbol{\delta}\|_{\infty}\le \epsilon$, to mislead victim VLMs. The pipeline commences by generating a set of visually grounded anchors from the target text $T_{\mathrm{tgt}}$ via a frozen Text-to-Image (T2I) model. Subsequently, we perform iterative optimization on a CLIP surrogate. In each step, we extract hierarchical features from the perturbed image. These features are then aligned with the weighted anchors and synchronized with the target text embeddings across multiple depths. The perturbation $\boldsymbol{\delta}$ is updated via gradient backpropagation to minimize the total loss, yielding the final transferable adversarial image. In the following subsections, we will describe the three key components of this framework.

\section{Methodology}
\label{sec:method}
As illustrated in Fig.~\ref{fig:attack_total}, we present SGHA-Attack, a transfer-based targeted attack framework designed to craft an adversarial example $\boldsymbol{x}_{\mathrm{adv}}$ by injecting a human-imperceptible perturbation $\boldsymbol{\delta}$ into a clean image $\boldsymbol{x}$ (i.e., $\boldsymbol{x}_{\mathrm{adv}} = \boldsymbol{x} + \boldsymbol{\delta}$ subject to $\|\boldsymbol{\delta}\|_{\infty}\le \epsilon$), aiming to mislead victim VLMs into generating attacker-specified outputs. 
The pipeline commences by constructing a visually grounded {reference pool} $\mathcal{X}_{ref}$ via a frozen Text-to-Image (T2I) model conditioned on the target text $T_{\mathrm{tgt}}$, from which the Top-$K$ most semantically relevant instances are adaptively selected as anchors.
Subsequently, we perform iterative optimization on a CLIP surrogate, where the perturbation $\boldsymbol{\delta}$ is updated via gradient backpropagation to strictly align the hierarchical features of $\boldsymbol{x}_{\mathrm{adv}}$ with these weighted anchor references while synchronizing visual and textual tokens across multiple depths, ultimately yielding the final transferable adversarial image.
In the following subsections, we systematically elaborate on the three synergistic components that constitute this framework.

\subsection{Semantic-Guided Anchor Injection}
\label{subsec:anchor}

AttackVLM~\cite{AttackVLM} shows that transfer-based targeted attacks can be driven by either text--image matching or image--image matching on a surrogate model. 
For text--image matching, we align the surrogate embedding of the adversarial image with the target text embedding. 
Let $\phi(\cdot)$ and $\psi(\cdot)$ denote the surrogate image and text encoders, respectively, and define cosine distance as $\mathcal{D}(\mathbf{a},\mathbf{b})=1-\cos(\mathbf{a},\mathbf{b})$. 
Here, $\phi(\cdot)$ produces a {global} image embedding used for the following matching objectives.
A standard objective is
\begin{equation}
\label{eq:ltxt}
    \mathcal{L}_{txt} = \mathcal{D}\big(\phi(\boldsymbol{x}_{adv}), \psi(T_{tgt})\big).
\end{equation}

For image--image matching, the adversarial image is guided toward a reference image by matching their surrogate embeddings, which provides more visually grounded supervision. 
However, existing methods\cite{AttackVLM,COA,M-Attack,AdvDiffVLM} often rely on a {single} reference instance, leading to narrow guidance: one image may capture only one appearance of the target concept, and the optimization can become sensitive to that particular target. 
This sensitivity makes targeted transfer less reliable when the victim model differs in backbone, alignment module, or processing pipeline.

To address this single-target limitation, we propose SGAI.
Instead of using one target reference, SGAI constructs a set of {semantic anchors} that represent diverse visual realizations of the target concept, and uses them jointly as guidance via a weighted mixture.
Concretely, we first build a reference pool $\mathcal{X}_{ref}$ by sampling from a text-to-image generator conditioned on $T_{tgt}$ (e.g., a diffusion model).
Given the target text embedding $\psi(T_{tgt})$, we select the Top-$K$ anchors according to cosine similarity under the surrogate:
\begin{equation}
\label{eq:topk}
    \{\boldsymbol{x}_{anc}^k\}_{k=1}^{K} =
    \mathrm{TopK}_{\boldsymbol{x}\in\mathcal{X}_{ref}}
    \cos\big(\phi(\boldsymbol{x}),\psi(T_{tgt})\big).
\end{equation}
We then compute importance weights using a temperature-scaled softmax:
\begin{equation}
\label{eq:weights}
    w_k=
    \frac{\exp\!\left(\cos(\phi(\boldsymbol{x}_{anc}^k),\psi(T_{tgt}))/\tau\right)}
    {\sum_{j=1}^{K}\exp\!\left(\cos(\phi(\boldsymbol{x}_{anc}^j),\psi(T_{tgt}))/\tau\right)}.
\end{equation}
Finally, we define an anchor-guided objective that aligns the adversarial embedding with a weighted mixture of anchor embeddings:
\begin{equation}
\label{eq:lanc}
    \mathcal{L}_{anc}=
    1-\sum_{k=1}^{K} w_k\cdot \cos\!\big(\phi(\boldsymbol{x}_{adv}),\phi(\boldsymbol{x}_{anc}^k)\big).
\end{equation}

Compared with a single reference, the weighted anchor mixture provides richer and more stable guidance: different anchors cover different appearances of the same target concept, while the weights automatically emphasize the most relevant anchors. 
In practice, this reduces sensitivity to a single target and improves targeted transfer across diverse victim VLMs.

From an efficiency perspective, SGAI introduces limited overhead. 
Anchor features $\phi(\boldsymbol{x}_{anc}^k)$ can be computed once and detached, and the iterative optimization only requires computing $\phi(\boldsymbol{x}_{adv})$ each iteration. 
We report the runtime in Table~\ref{tab:runtime}.
%Thus, the additional cost mainly comes from generating $\mathcal{X}_{ref}$ and the one-time ranking/selection in Eq.~\eqref{eq:topk}. 

\subsection{Hierarchical Visual Structure Alignment}
\label{subsec:visual}

SGAI provides a robust target guidance at the global embedding level, but many transfer-based attacks still impose constraints only on the final-layer embedding of the surrogate visual encoder. 
For ViT encoders, representations evolve progressively across layers: shallow layers capture low-level patterns, while deeper layers encode higher-level semantics. 
Consequently, restricting supervision to the final embedding may underutilize intermediate representations that influence transferability, especially when victim models vary in depth, patch size, or projector design. 
To address this, we propose HVSA, which aligns intermediate visual representations of $\boldsymbol{x}_{adv}$ to anchor-derived targets at multiple layers.

Let $\mathcal{L}$ denote a set of selected transformer layers.
Denote by $\phi^l(\cdot)$ the intermediate feature map extracted at layer $l\in\mathcal{L}$. 
Using the anchor weights $\{w_k\}$ from Sec.~\ref{subsec:anchor}, we construct a layer-wise weighted anchor target:
\begin{equation}
\label{eq:Ftgt}
    \hat{\mathbf{F}}^{l}_{anc}=\sum_{k=1}^{K} w_k\cdot \phi^{l}(\boldsymbol{x}_{anc}^k).
\end{equation}
This anchor-derived target provides a clear intermediate-layer reference based on the anchor images, making the supervision at each layer well-defined.

We further distinguish the roles of different ViT tokens. 
Let $\mathbf{F}^{l}_{adv}=\phi^l(\boldsymbol{x}_{adv})\in\mathbb{R}^{(N+1)\times D}$ be the feature map at layer $l$, where the first token corresponds to CLS and the remaining $N$ tokens correspond to spatial patches. 
We decompose $\mathbf{F}^{l}_{adv}$ and $\hat{\mathbf{F}}^{l}_{anc}$ into global tokens
$(\mathbf{f}^{l}_{cls},\hat{\mathbf{f}}^{l}_{cls})\in\mathbb{R}^{D}$ and spatial tokens
$(\mathbf{F}^{l}_{spa},\hat{\mathbf{F}}^{l}_{spa})\in\mathbb{R}^{N\times D}$. 
We mean-pool patch tokens to obtain a single spatial descriptor per layer:
$\overline{\mathbf{F}}^{l}_{spa}=\frac{1}{N}\sum_{i=1}^{N}\mathbf{F}^{l}_{spa}[i]$ (and similarly for $\overline{\hat{\mathbf{F}}}^{l}_{spa}$). 
Mean pooling provides a compact spatial summary without requiring strict patch-to-patch correspondence, making the alignment more robust to differences in patch size and common preprocessing (e.g., resizing or cropping) across models.
The HVSA loss is then defined as:
\begin{equation}
\label{eq:lfeat}
\small
    \mathcal{L}_{feat}
    =\sum_{l\in\mathcal{L}}
    \Big(
        \lambda_{cls}\,\mathcal{D}(\mathbf{f}^{l}_{cls},\hat{\mathbf{f}}^{l}_{cls})
        +\lambda_{spa}\,\mathcal{D}(\overline{\mathbf{F}}^{l}_{spa},\overline{\hat{\mathbf{F}}}^{l}_{spa})
    \Big).
\end{equation}

The term weighted by $\lambda_{cls}$ aligns the CLS token, encouraging the adversarial example to match the target at a global, image-level semantics across selected layers. 
The term weighted by $\lambda_{spa}$ aligns the pooled patch tokens, encouraging consistency in spatial cues related to the target. 
We separate these two terms to provide complementary supervision: the CLS term focuses on global semantics, while the spatial term preserves coarse spatial cues, leading to a more balanced alignment signal.

\subsection{Cross-Modal Latent Space Synchronization}
\label{subsec:cross_modal}

SGAI and HVSA mainly constrain the visual stream: SGAI provides target-guided anchors, and HVSA aligns intermediate visual features to anchor-derived targets across layers. 
However, a VLM produces outputs based on cross-modal interactions between vision and language. 
Thus, beyond final-layer matching, we enforce intermediate cross-modal alignment so that the intermediate visual features of $\boldsymbol{x}_{adv}$ are aligned with the target text features.

A key challenge is that intermediate visual and textual features are not directly comparable before projection, since they may differ in dimension and representation space. 
To make them comparable, we map intermediate features into a shared latent subspace using projection heads available in the surrogate. 
For each layer $l\in\mathcal{L}$, we extract the intermediate CLS visual token $\mathbf{f}^{l}_{cls}$ from the surrogate image encoder, and an intermediate text feature $\mathbf{t}^{l}_{eos}$ from the surrogate text encoder, taken from the EOS token at the corresponding depth. 

We then apply the visual and text projections and minimize the distance between them:
\begin{equation}
\label{eq:lmid}
    \mathcal{L}_{mid}
    =\sum_{l\in\mathcal{L}}
    \mathcal{D}\Big(
        \mathbf{f}^{l}_{cls}\mathbf{W}_{proj},
        \mathbf{t}^{l}_{eos}\mathbf{P}_{txt}
    \Big),
\end{equation}
where $\mathbf{W}_{proj}$ and $\mathbf{P}_{txt}$ are the surrogate's original visual and textual projection heads used for image--text alignment, mapping features into a shared latent subspace.

CLSS imposes cross-modal alignment at intermediate layers: after projection into the shared latent subspace, intermediate adversarial visual features are aligned with the target text features, instead of relying only on final-layer matching. 
During optimization, $\mathcal{L}_{mid}$ provides cross-modal gradients at multiple depths, which improves targeted transfer when victim VLMs use different projection modules or cross-modal alignment schemes.

\begin{algorithm}[t]
\caption{The overall algorithm of SGHA-Attack}
\label{alg:attack}
\begin{algorithmic}[1]
\REQUIRE Clean image $\boldsymbol{x}$, target text $T_{tgt}$, text-to-image model $\mathrm{T2I}$, surrogate image encoder $\phi$, text encoder $\psi$, visual projection $\mathbf{W}_{proj}$, text projection $\mathbf{P}_{txt}$, perturbation budget $\epsilon$, step size $\alpha$, steps $N$, layer set $\mathcal{L}$, weights $\lambda_{anc},\lambda_{feat},\lambda_{cls},\lambda_{spa},\lambda_{mid}$, temperature $\tau$.
\ENSURE Adversarial image $\boldsymbol{x}_{adv}$.

\vspace{0.1cm}
\item[] \textbf{Phase 1: Pre-processing}
\STATE Generate reference pool $\mathcal{X}_{ref} \leftarrow \mathrm{T2I}(T_{tgt})$
\STATE Select Top-$K$ anchors $\{\boldsymbol{x}_{anc}^k\}_{k=1}^{K}\subset\mathcal{X}_{ref}$ by Eq.~\eqref{eq:topk}
\STATE Compute anchor weights $w_k$ via Eq.~\eqref{eq:weights} with temperature $\tau$, forming a softmax mixture \hfill \textcolor{gray}{// SGAI}
\FOR{each selected layer $l \in \mathcal{L}$}
    \item[] \textcolor{gray}{// Extract weighted anchor features}
    \STATE $\hat{\mathbf{f}}^{l}_{cls}, \overline{\hat{\mathbf{F}}}^{l}_{spa}
    \leftarrow \sum_{k=1}^{K} w_k\cdot \phi^{l}(\boldsymbol{x}_{anc}^k)$
    \item[] \textcolor{gray}{// Extract target text features}
    \STATE $\mathbf{t}^{l}_{eos}\leftarrow \psi^{l}(T_{tgt})$
\ENDFOR

\vspace{0.1cm}
\item[] \textbf{Phase 2: Adversarial Optimization}
\STATE Initialize perturbation $\boldsymbol{\delta}^{0} \leftarrow \mathbf{0}$
\FOR{$t=0$ \TO $N-1$}
    \STATE $\boldsymbol{x}^{t}_{adv} \leftarrow \boldsymbol{x} + \boldsymbol{\delta}^{t}$
    \STATE $\mathcal{L}_{txt}\leftarrow \mathcal{D}\big(\phi(\boldsymbol{x}^{t}_{adv}),\psi(T_{tgt})\big)$ \hfill \textcolor{gray}{// Final embedding}
    \STATE $\mathcal{L}_{anc}\leftarrow 1-\sum_{k=1}^{K} w_k\cdot \cos\big(\phi(\boldsymbol{x}^{t}_{adv}),\phi(\boldsymbol{x}_{anc}^k)\big)$

    \STATE $\mathcal{L}_{feat}\leftarrow 0, \mathcal{L}_{mid}\leftarrow 0$
    \FOR{each selected layer $l \in \mathcal{L}$}
        \item[] \textcolor{gray}{// Extract adversarial features}
        \STATE $\mathbf{f}^{l}_{cls}, \overline{\mathbf{F}}^{l}_{spa} \leftarrow \phi^{l}(\boldsymbol{x}^{t}_{adv})$
        \item[] \textcolor{gray}{// HVSA: Align intermediate visual features}
        \STATE $\mathcal{L}_{feat}\leftarrow \mathcal{L}_{feat}
            +\lambda_{cls}\mathcal{D}(\mathbf{f}^{l}_{cls},\hat{\mathbf{f}}^{l}_{cls})
            +\lambda_{spa}\mathcal{D}(\overline{\mathbf{F}}^{l}_{spa},\overline{\hat{\mathbf{F}}}^{l}_{spa})$  
        \item[]   \textcolor{gray}{// CLSS: Synchronize cross-modal features}
        \STATE $\mathcal{L}_{mid}\leftarrow \mathcal{L}_{mid}
            +\mathcal{D}\big(\mathbf{f}^{l}_{cls}\mathbf{W}_{proj},\mathbf{t}^{l}_{eos}\mathbf{P}_{txt}\big)$  
    \ENDFOR
    \STATE $\mathcal{L}_{total}\leftarrow \mathcal{L}_{txt}+\lambda_{anc}\mathcal{L}_{anc}+\lambda_{feat}\mathcal{L}_{feat}+\lambda_{mid}\mathcal{L}_{mid}$
    \STATE $\boldsymbol{g}^{t}\leftarrow \nabla_{\boldsymbol{x}^{t}_{adv}}\mathcal{L}_{total}$ \hfill \textcolor{gray}{// Gradient backpropagation}
    \STATE $\boldsymbol{\delta}^{t+1}\leftarrow \mathrm{Clip}_{[-\epsilon, \epsilon]}\big(\boldsymbol{\delta}^{t} - \alpha\cdot \mathrm{sign}(\boldsymbol{g}^{t})\big)$ \hfill \textcolor{gray}{// Update}
\ENDFOR
\STATE $\boldsymbol{x}_{adv}\leftarrow \mathrm{Clip}_{[-\epsilon, \epsilon]}(\boldsymbol{x} + \boldsymbol{\delta}^{N})$
\RETURN $\boldsymbol{x}_{adv}$
\end{algorithmic}
\end{algorithm}

\subsection{Total Objective and Optimization Procedure}
\label{subsec:algorithm}

We integrate the three modules, SGAI, HVSA, and CLSS, into a unified objective function:
\begin{equation}
\label{eq:ltotal}
    \mathcal{L}_{total}
    =\mathcal{L}_{txt}
    +\lambda_{anc}\mathcal{L}_{anc}
    +\lambda_{feat}\mathcal{L}_{feat}
    +\lambda_{mid}\mathcal{L}_{mid},
\end{equation}
where $\lambda_{anc}$, $\lambda_{feat}$, and $\lambda_{mid}$ balance anchor guidance, hierarchical visual alignment, and intermediate cross-modal synchronization. 
We optimize $\boldsymbol{x}_{adv}$ using projected gradient descent under an $\ell_\infty$ budget $\epsilon$. 
At each iteration, the adversarial image is updated using the sign of the gradient and projected back to the feasible $\ell_\infty$ ball around $\boldsymbol{x}$, pixel values are also clipped to the valid range.
Algorithm~\ref{alg:attack} summarizes the complete procedure.

\section{Experiments}
\subsection{Experimental Settings}
\noindent \textbf{Datasets.} We follow the evaluation protocol in~\cite{AttackVLM,COA}. Specifically, we sample 1,000 images from ImageNet-1K~\cite{ImageNet} as clean inputs and sample 1,000 text descriptions from MS-COCO~\cite{COCO} as target prompts. Each ImageNet image is randomly paired with one MS-COCO description, which typically yields a semantic mismatch between the source image and the target text, forming a challenging targeted setting. Following~\cite{AttackVLM}, we further use Stable Diffusion~\cite{StableDiffusion} to synthesize text-conditioned target reference images from the target descriptions.

\begin{table}[t]
    \centering
    %\caption{Configurations of the victim VLMs evaluated in this study.} 
    \caption{Configurations of the victim VLMs.} 
    \label{tab:victim_vlms}
    \renewcommand{\arraystretch}{1.2}
    \setlength{\tabcolsep}{5pt}
    \resizebox{\linewidth}{!}{
    \begin{tabular}{llccc}
        \toprule
        \textbf{Category} & \textbf{Model} & \textbf{Params} & \textbf{Vision Module} & \textbf{Text Module / LLM} \\
        \midrule
        \multirow{6}{*}{\textbf{Open-Source}} 
        & UniDiffuser~\cite{Unidiffuser} & 1.4B & CLIP ViT-B/32 & UViT-H (Diffusion) \\
        & BLIP-2~\cite{BLIP2} & 12.1B & ViT-G/14 (EVA-CLIP) & FLAN-T5 XXL \\
        & InstructBLIP~\cite{InstructBLIP} & 14.2B & ViT-G/14 (EVA-CLIP) & Vicuna-13B \\
        & MiniGPT-4~\cite{Minigpt-4} & 14.2B & ViT-G/14 (EVA-CLIP) & Vicuna-13B \\
        & LLaVA~\cite{LLaVA} & 13.3B & CLIP ViT-L/14 & Vicuna-13B \\
        & LLaVA-NeXT~\cite{llavanext} & 72.3B & CLIP ViT-L/14-336px & Qwen1.5-72B-Chat \\
        \midrule
        \multirow{3}{*}{\textbf{Closed-Source}} 
        & OpenAI GPT-4o & \multicolumn{3}{c}{Undisclosed} \\ 
        & Google Gemini-2.0 & \multicolumn{3}{c}{Undisclosed} \\
        & Anthropic Claude-3.5 & \multicolumn{3}{c}{Undisclosed} \\
        \bottomrule
    \end{tabular}
    }
\end{table}

\noindent \textbf{Victim VLMs.} We select a diverse set of victim models to evaluate the effectiveness of our attack, including six state-of-the-art open-source VLMs and three leading closed-source commercial models. The detailed configurations are summarized in Table \ref{tab:victim_vlms}. It is worth noting that certain models (e.g., MiniGPT-4 and InstructBLIP) share identical vision backbones and parameter scales but employ distinct cross-modal alignment mechanisms (e.g., linear projection vs. Q-Former), thereby serving as diverse testbeds for evaluating transferability. Regarding the black-box attack setting, we follow the attack protocols in~\cite{AttackVLM} and~\cite{M-Attack} for open-source and closed-source commercial models, respectively.

\noindent \textbf{Baselines.} We benchmark against state-of-the-art transfer-based targeted attacks categorized into two groups: VLM-specific approaches, including MF-it~\cite{AttackVLM}, MF-ii~\cite{AttackVLM}, COA~\cite{COA}, AdvDiffVLM~\cite{AdvDiffVLM}, and M-Attack~\cite{M-Attack}. Following~\cite{AdvDiffVLM}, we also adapt representative SOTA image classification methods BSR~\cite{BSR} and OPS~\cite{OPS} for the VLM task by replacing their classification loss with a target-embedding alignment loss.

\begin{table*}[!htbp]
\centering
\caption{Quantitative comparison of transfer-based targeted attacks against black-box VLMs. We report the CLIP Score ($\uparrow$) to measure semantic consistency across various text encoders and their ensemble average, alongside the LLM-based ASR evaluated by GPT-4. In the following tables, highlighted, underlined, and gray-shaded values indicate the best, second-best, and our method’s results, respectively, for each case.}
%The best results are highlighted in \textbf{bold}, the second best are \underline{underlined}, and the gray shading indicates our proposed method.
\label{tab:main_results}
% === 样式设置 ===
\renewcommand{\arraystretch}{0.87} % 保持紧凑的行高
%\setlength{\tabcolsep}{2.8pt}     % 保持紧凑的列间距
%\footnotesize                     % 保持小字号
% ================
\begin{tabular}{ll|cccccc|cc}
\toprule
\multirow{2}{*}{\textbf{VLM Model}} & \multirow{2}{*}{\textbf{Attack Method}} & \multicolumn{6}{c|}{\textbf{Text Encoder for Evaluation (CLIP Score $\uparrow$)}} & \multicolumn{2}{c}{\textbf{ASR ($\uparrow$)}} \\
 &  & RN50 & RN101 & ViT-B/16 & ViT-B/32 & ViT-B/14 & Ensemble & $\text{ASR}_{\text{fool}}$ & $\text{ASR}_{\text{target}}$ \\ \midrule

% === UniDiffuser ===
\multirow{9}{*}{\shortstack[l]{\textbf{UniDiffuser} \cite{Unidiffuser}}} 
 & Clean Image & 0.4417 & 0.4275 & 0.4504 & 0.4690 & 0.3215 & 0.4220 & 0.00\% & 0.00\% \\
 & BSR \cite{BSR} & 0.5415 & 0.5244 & 0.5542 & 0.5728 & 0.4382 & 0.5262 & 72.30\% & 18.00\% \\
 & OPS \cite{OPS} & 0.6470 & 0.6284 & 0.6592 & 0.6748 & 0.5635 & 0.6345 & 94.40\% & 52.30\% \\
 & M-Attack \cite{M-Attack} & 0.5620 & 0.5454 & 0.5747 & 0.5908 & 0.4617 & 0.5469 & 73.30\% & 23.10\% \\
 & AdvDiffVLM \cite{AdvDiffVLM} &0.6465 &0.6289 &0.6582 &0.6748 &0.5630 &0.6343 & 90.20\% & 51.90\%  \\
 & MF-it \cite{AttackVLM} & 0.6768 & 0.6577 & 0.6909 & 0.7065 & 0.5933 & 0.6650 & 91.10\% & 48.20\% \\
 & MF-ii \cite{AttackVLM} & 0.7061 & \underline{0.6929} & 0.7188 & 0.7324 & \underline{0.6353} & 0.6970 & 98.00\% & 72.10\% \\
 & COA \cite{COA} & \underline{0.7100} & 0.6880 & \underline{0.7217} & \underline{0.7398} & 0.6284 & \underline{0.6976} & \underline{98.80\%} & \underline{73.20\%} \\
 \rowcolor{Gray} \cellcolor{white} & \textbf{Ours} & \textbf{0.7793} & \textbf{0.7647} & \textbf{0.7891} & \textbf{0.7998} & \textbf{0.7197} & \textbf{0.7705} & \textbf{99.80\%} & \textbf{89.40\%} \\ \midrule

% === BLIP2 ===
\multirow{9}{*}{\shortstack[l]{\textbf{BLIP2} \cite{BLIP2} }} 
 & Clean Image & 0.4663 & 0.4526 & 0.4763 & 0.4954 & 0.3491 & 0.4480 & 0.00\% & 0.00\% \\
 & BSR \cite{BSR} & 0.5874 & 0.5659 & 0.5991 & 0.6162 & 0.4858 & 0.5709 & 55.00\% & 20.80\% \\
 & OPS \cite{OPS} & 0.6475 & 0.6255 & 0.6582 & 0.6743 & 0.5581 & 0.6327 & 72.10\% & 34.80\% \\
 & M-Attack \cite{M-Attack} & 0.5376 & 0.5161 & 0.5498 & 0.5669 & 0.4292 & 0.5199 & 30.80\% & 9.90\% \\
 & AdvDiffVLM \cite{AdvDiffVLM} &0.6333 &0.6133 &0.6450 &0.6616 &0.5440 &0.6194 & 63.70\% & 32.90\% \\
 & MF-it \cite{AttackVLM} & 0.7607 & 0.7446 & 0.7715 & 0.7813 & 0.6943 & 0.7505 & 93.50\% & 70.60\% \\
 & MF-ii \cite{AttackVLM} & 0.7676 & 0.7529 & 0.7773 & 0.7896 & 0.7041 & 0.7583 & 97.80\% & \underline{82.80\%} \\
 & COA \cite{COA} & \underline{0.7754} & \underline{0.7598} & \underline{0.7852} & \underline{0.7959} & \underline{0.7110} & \underline{0.7654} & \underline{99.30\%} & 80.20\% \\
 \rowcolor{Gray} \cellcolor{white} & \textbf{Ours} & \textbf{0.8359} & \textbf{0.8228} & \textbf{0.8438} & \textbf{0.8525} & \textbf{0.7861} & \textbf{0.8282} & \textbf{99.40\%} & \textbf{94.70\%} \\ \midrule

% === InstructBLIP ===
\multirow{9}{*}{\shortstack[l]{\textbf{InstructBLIP} \cite{InstructBLIP}}} 
 & Clean Image & 0.4639 & 0.4544 & 0.4756 & 0.4934 & 0.3442 & 0.4463 & 0.00\% & 0.00\% \\
 & BSR \cite{BSR} & 0.5801 & 0.5635 & 0.5903 & 0.6060 & 0.4775 & 0.5635 & 55.20\% & 22.00\% \\
 & OPS \cite{OPS} & 0.6367 & 0.6196 & 0.6484 & 0.6626 & 0.5444 & 0.6224 & 71.70\% & 34.10\% \\
 & M-Attack \cite{M-Attack} & 0.5337 & 0.5161 & 0.5435 & 0.5601 & 0.4209 & 0.5148 & 32.00\% & 9.30\% \\
 & AdvDiffVLM \cite{AdvDiffVLM} &0.6333 &0.6167 &0.6455 &0.6587 &0.5405 &0.6189 & 66.80\% & 35.20\%  \\
 & MF-it \cite{AttackVLM} & 0.7603 & 0.7456 & 0.7700 & 0.7803 & 0.6948 & 0.7502 & 94.10\% & 73.70\% \\
 & MF-ii \cite{AttackVLM} & 0.7695 & 0.7544 & 0.7783 & 0.7886 & 0.7036 & 0.7589 & 98.40\% & \underline{84.00\%} \\
 & COA \cite{COA} & \underline{0.7720} & \underline{0.7583} & \underline{0.7832} & \underline{0.7920} & \underline{0.7075} & \underline{0.7626} & \underline{99.10\%} & 81.30\% \\
 \rowcolor{Gray} \cellcolor{white} & \textbf{Ours} & \textbf{0.8345} & \textbf{0.8223} & \textbf{0.8423} & \textbf{0.8501} & \textbf{0.7827} & \textbf{0.8264} & \textbf{99.30\%} & \textbf{95.70\%} \\ \midrule

% === MiniGPT-4 ===
\multirow{9}{*}{\shortstack[l]{\textbf{MiniGPT-4} \cite{Minigpt-4} }} 
 & Clean Image & 0.2871 & 0.3916 & 0.3252 & 0.3569 & 0.2131 & 0.3148 & 0.00\% & 0.00\% \\
 & BSR \cite{BSR} & 0.3696 & 0.4873 & 0.4229 & 0.4458 & 0.3196 & 0.4090 & 55.20\% & 20.30\% \\
 & OPS \cite{OPS} & 0.3945 & 0.5288 & 0.4536 & 0.4724 & 0.3596 & 0.4418 & 71.80\% & 35.60\% \\
 & M-Attack \cite{M-Attack} & 0.3311 & 0.4446 & 0.3792 & 0.4063 & 0.2703 & 0.3663 & 29.90\% & 7.90\% \\
 & AdvDiffVLM \cite{AdvDiffVLM} &0.3982 &0.5269 &0.4561 &0.4773 &0.3630 &0.4443 & 63.70\% & 31.50\% \\
 & MF-it \cite{AttackVLM} & 0.4929 & 0.6270 & 0.5591 & 0.5737 & 0.4793 & 0.5464 & 94.60\% & 72.00\% \\
 & MF-ii \cite{AttackVLM} & 0.5112 & 0.6431 & 0.5781 & 0.5898 & 0.4998 & 0.5644 & 98.10\% & \underline{81.00\%} \\
 & COA \cite{COA} & \underline{0.5161} & \underline{0.6445} & \underline{0.5850} & \underline{0.5962} & \underline{0.5054} & \underline{0.5694} & \underline{99.60\%} & 78.70\% \\
 \rowcolor{Gray} \cellcolor{white} & \textbf{Ours} & \textbf{0.5850} & \textbf{0.7061} & \textbf{0.6519} & \textbf{0.6602} & \textbf{0.5801} & \textbf{0.6366} & \textbf{99.70\%} & \textbf{95.60\%} \\ \midrule

% === LLaVA ===
\multirow{9}{*}{\shortstack[l]{\textbf{LLaVA} \cite{LLaVA} }} 
 & Clean Image & 0.3413 & 0.4395 & 0.3770 & 0.4021 & 0.2341 & 0.3588 & 0.00\% & 0.00\% \\
 & BSR \cite{BSR} & 0.4734 & 0.5444 & 0.5171 & 0.5308 & 0.3901 & 0.4912 & 64.80\% & 25.30\% \\
 & OPS \cite{OPS} & 0.5064 & 0.5767 & 0.5503 & 0.5635 & 0.4277 & 0.5249 & 72.50\% & 39.70\% \\
 & M-Attack \cite{M-Attack} & 0.4233 & 0.5015 & 0.4609 & 0.4802 & 0.3252 & 0.4382 & 40.40\% & 11.10\% \\
 & AdvDiffVLM \cite{AdvDiffVLM} &0.4800 &0.5513 &0.5205 &0.5396 &0.3945 &0.4972 & 60.70\% & 25.40\% \\
 & MF-it \cite{AttackVLM} & 0.4922 & 0.5640 & 0.5366 & 0.5523 & 0.4209 & 0.5132 & 67.70\% & 27.60\% \\
 & MF-ii \cite{AttackVLM} & \underline{0.5625} & \underline{0.6274} & \underline{0.6128} & \underline{0.6211} & \underline{0.5000} & \underline{0.5848} & 88.00\% & \underline{61.00\%} \\
 & COA \cite{COA} & 0.5200 & 0.5913 & 0.5679 & 0.5786 & 0.4495 & 0.5415 & \underline{88.90\%} & 42.40\% \\
 \rowcolor{Gray} \cellcolor{white} & \textbf{Ours} & \textbf{0.6821} & \textbf{0.7246} & \textbf{0.7358} & \textbf{0.7324} & \textbf{0.6460} & \textbf{0.7042} & \textbf{99.10\%} & \textbf{96.40\%} \\ \midrule

% === LLaVA-NeXT ===
\multirow{9}{*}{\shortstack[l]{\textbf{LLaVA-NeXT} \cite{llavanext} }} 
 & Clean Image & 0.2673 & 0.3809 & 0.3054 & 0.3120 & 0.1897 & 0.2911 & 0.00\% & 0.00\% \\
 & BSR \cite{BSR} & 0.3723 & 0.4729 & \underline{0.4143} & 0.4097 & 0.2869 & 0.3912 & 56.50\% & 14.80\% \\
 & OPS \cite{OPS} & 0.3662 & 0.4727 & 0.4067 & 0.4019 & 0.2791 & 0.3853 & 53.60\% & 14.90\% \\
 & M-Attack \cite{M-Attack} & 0.3181 & 0.4268 & 0.3567 & 0.3562 & 0.2300 & 0.3375 & 29.80\% & 4.40\% \\
 & AdvDiffVLM \cite{AdvDiffVLM} &0.3306 &0.4382 &0.3706 &0.3713 &0.2441 &0.3510 & 37.90\% & 5.80\%  \\
 & MF-it \cite{AttackVLM} & 0.3325 & 0.4377 & 0.3674 & 0.3711 & 0.2417 & 0.3501 & 33.70\% & 4.10\% \\
 & MF-ii \cite{AttackVLM} & \underline{0.3738} & \underline{0.4780} & 0.4104 & \underline{0.4104} & \underline{0.2874} & \underline{0.3920} & 49.90\% & \underline{15.60\%} \\
 & COA \cite{COA} & 0.3650 & 0.4648 & 0.4031 & 0.4019 & 0.2703 & 0.3810 & \underline{60.30\%} & 10.60\% \\
 \rowcolor{Gray} \cellcolor{white} & \textbf{Ours} & \textbf{0.5391} & \textbf{0.6338} & \textbf{0.5869} & \textbf{0.5835} & \textbf{0.4944} & \textbf{0.5675} & \textbf{92.60\%} & \textbf{73.40\%} \\ 
\bottomrule
\end{tabular}%
\end{table*}

\noindent \textbf{Evaluation metrics.} Following standard protocols~\cite{AttackVLM,COA}, we employ CLIP-Score~\cite{AttackVLM} to measure semantic consistency and the LLM-based (GPT-4) Attack Success Rate (ASR)~\cite{COA} to assess effectiveness, specifically utilizing $\text{ASR}_{\text{fool}}$ for {untargeted attack success} and $\text{ASR}_{\text{target}}$ for {targeted attack success}. Additionally, we utilize Structural Similarity (SSIM)~\cite{ssim} and Peak Signal-to-Noise Ratio (PSNR) to evaluate the visual imperceptibility of the generated adversarial examples.

\noindent \textbf{Implementation Details.}\footnote{Code available at: \url{https://github.com/BiiiGerrr/SGHA-Attack}} We utilize PGD~\cite{pgd} with a maximum perturbation of $\epsilon=8/255$ and 100 iterations under the $L_\infty$ norm across all methods to ensure fair comparison. For SGHA-Attack, we set anchor parameters to $K=5$, $\tau=5$, and $\lambda_{anc}=1$. To determine the intermediate layers $\mathcal{L}$, we adopt a {Deep-Layer Uniform Sampling} strategy, specifically selecting layers $\{7, 9, 11\}$ for the base ViT-B/32 model. The balancing weights $\{\lambda_{feat}, \lambda_{cls}, \lambda_{spa}, \lambda_{mid}\}$ are scaled according to the backbone architecture, for instance, they are set to {1.5, 1.0, 0.7, 2.5} for ViT-B/32. For larger models like ViT-L and ViT-G, we proportionally increase the feature alignment weights and apply the corresponding layer mappings to accommodate their higher-dimensional feature spaces and deeper architectures.

\begin{figure*}[!ht]
    \centering
    \includegraphics[width=\linewidth]{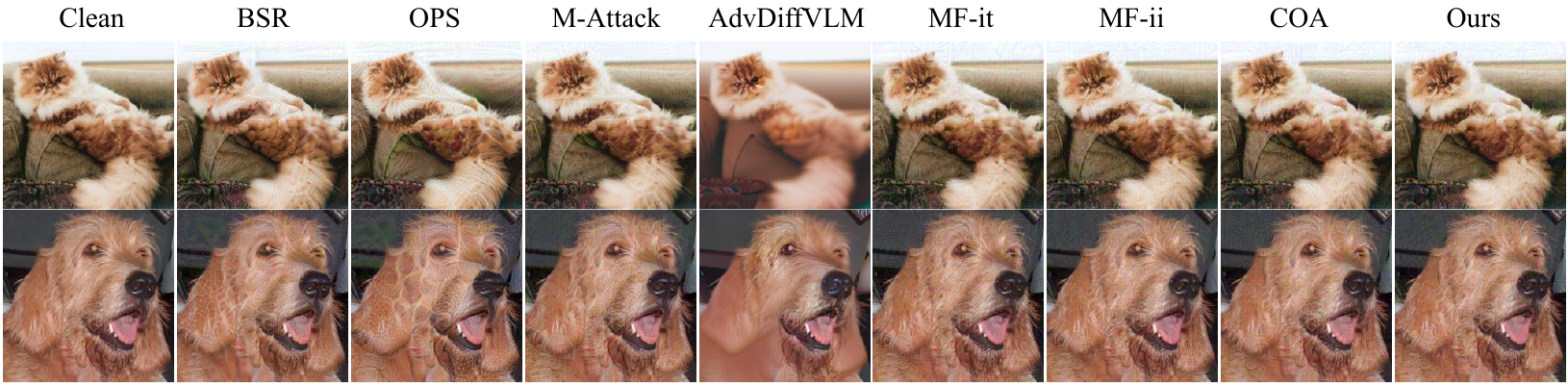} % 请确认文件名
    \caption{Visual comparison of adversarial examples generated by various attack methods.}
    \label{fig:visual_comparison}
\end{figure*}

\subsection{Comprehensive Evaluation on Open-Source VLMs}
\label{sec:experiments}

In this section, we present a comprehensive evaluation of our proposed method against state-of-the-art transfer-based targeted attacks. We conduct experiments across six representative open-source VLMs ranging from 1.4B to 72.3B parameters, including UniDiffuser, BLIP-2, InstructBLIP, MiniGPT-4, LLaVA, and LLaVA-NeXT. To ensure a consistent evaluation setting, we employ a unified prompt: ``\textit{What is the content of the image?}'' during the inference phase.

\noindent \textbf{Quantitative Attack Performance.}
We first conduct a quantitative comparison to assess the attack effectiveness. As reported in Table \ref{tab:main_results}, our method demonstrates superior performance across all metrics.

\begin{enumerate}
    \item Our method consistently achieves state-of-the-art performance in both $\text{ASR}_{\text{fool}}$ and $\text{ASR}_{\text{target}}$ metrics across all evaluated models. Specifically, regarding the challenging $\text{ASR}_{\text{target}}$ which measures the targeted semantic accuracy, SGHA-Attack demonstrates substantial improvements over the strongest baselines. For instance, on the 12.1B parameter BLIP-2 model, we achieve an $\text{ASR}_{\text{target}}$ of 94.70\%, significantly surpassing MF-ii (82.80\%) and COA (80.20\%). This indicates that our approach not only successfully misleads the VLMs but also precisely steers them toward the target semantics, reducing irrelevant outputs.

    \item In terms of semantic consistency, our approach yields significant gains in Ensemble CLIP Scores. High CLIP scores imply that the generated adversarial captions are semantically aligned with the target prompts across diverse vision encoders. As shown in Table \ref{tab:main_results}, SGHA-Attack obtains the highest scores on all six victim models (e.g., 0.8282 on BLIP-2 compared to 0.7654 for COA). This confirms that our hierarchical alignment generates generalizable semantic features rather than overfitting to surrogate-specific biases and artifacts, ensuring the adversarial perturbations remain effective across different feature extractors.

    \item Regarding robustness across model scales and architectures, our method exhibits strong generalization from the lightweight 1.4B UniDiffuser to the 72.3B LLaVA-NeXT. Regardless of the underlying cross-modal alignment mechanisms (e.g., linear projection or Q-Former), our attack maintains high efficacy. Notably, on LLaVA-NeXT where standard methods struggle due to stronger robustness, SGHA-Attack maintains a high success rate of 73.40\%. This validates that our multi-granularity strategy effectively generalizes across architectures defenses to establish robust semantic control.
\end{enumerate}

\begin{figure*}[!ht]
  \centering
  % width=\textwidth 表示图片宽度等于整个页面的文本宽度
  \includegraphics[width=\textwidth]{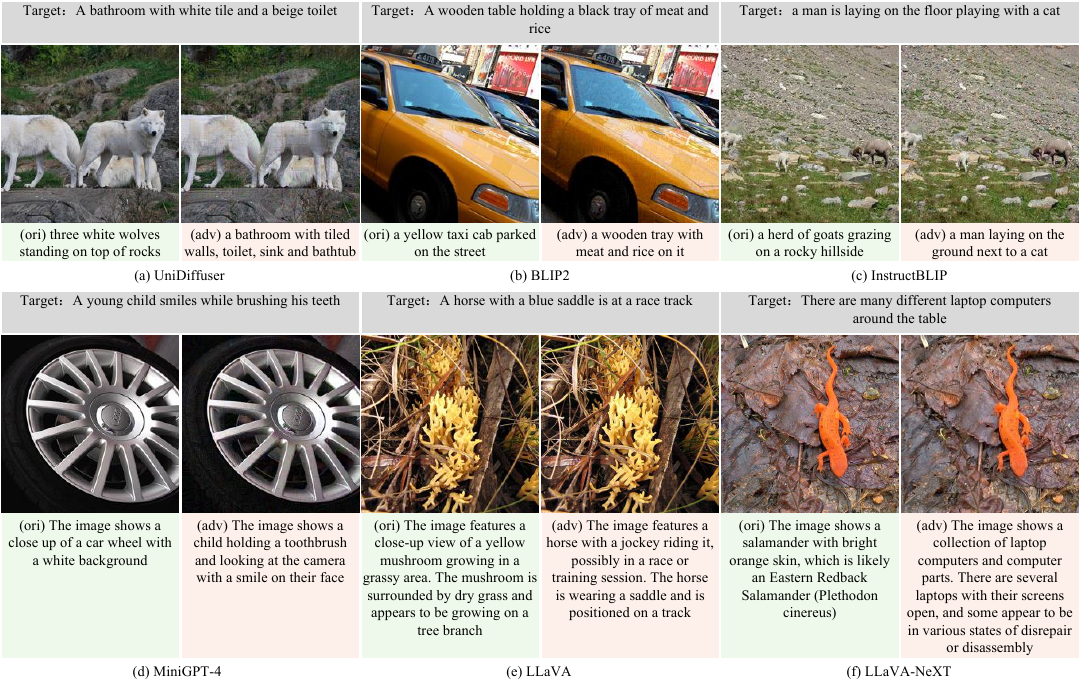}
  \caption{Qualitative results of our attack on multiple open-source VLMs. The target text prompt is shown above each image pair. For each example, the left image is the clean original with its caption, and the right image is the adversarial example generated by our method ($\epsilon=8/255$) with the model-generated caption. The adversarial captions closely align with the target texts, while the images remain visually similar to the originals.}

  \label{fig:vis_results}
\end{figure*}

\begin{table}[t]
\centering
\caption{Quantitative evaluation of visual quality.}
\label{tab:visual_metrics}
\renewcommand{\arraystretch}{1.1}
\setlength{\tabcolsep}{8pt} % 适当调整列间距以适应新增列
\footnotesize
%\resizebox{\linewidth}{!}{%
\begin{tabular}{c|l|cc}
\toprule
\textbf{Category} & \textbf{Method} & \textbf{SSIM} $\uparrow$ & \textbf{PSNR} $\uparrow$ \\
\midrule
% === Category 1: Unrestricted ===
Unrestricted & AdvDiffVLM~\cite{AdvDiffVLM} & 0.69 & 23.06 \\
\midrule
% === Category 2: Constrained ===
\multirow{7}{*}{\shortstack{Perturbation\\Constrained}}
 & BSR~\cite{BSR} & 0.83 & 31.65 \\
 & OPS~\cite{OPS} & 0.83 & 31.26 \\
 & M-Attack~\cite{M-Attack} & \underline{0.89} & 33.78 \\
 & MF-it~\cite{AttackVLM} & \underline{0.89} & \underline{33.98} \\
 & MF-ii~\cite{AttackVLM} & \textbf{0.90} & \textbf{34.28} \\
 & COA~\cite{COA} & \underline{0.89} & 33.68 \\
 & \cellcolor{Gray}\textbf{Ours} & \cellcolor{Gray}\underline{0.89} & \cellcolor{Gray}33.71 \\
\bottomrule
\end{tabular}
\end{table}

\noindent \textbf{Visual Quality Performance.}
We evaluate the visual quality and targeted effectiveness of the generated adversarial examples through both quantitative metrics reported in Table~\ref{tab:visual_metrics} and qualitative visualizations presented in Fig.~\ref{fig:visual_comparison} and Fig.~\ref{fig:vis_results}.

\begin{enumerate}
    \item As shown in Table~\ref{tab:visual_metrics}, performance varies significantly by attack category. Unrestricted methods like AdvDiffVLM suffer from noticeable structural degradation (SSIM 0.69) due to their diffusion-based generation process. In contrast, perturbation-constrained methods operate under a standard budget ($\epsilon=8/255$) and consistently maintain high {structural integrity}. Our method achieves an SSIM of 0.89 and PSNR of 33.71, demonstrating that our approach yields visual quality {comparable to state-of-the-art constrained baselines (e.g., MF-ii) while significantly surpassing unrestricted approaches}. This imperceptibility is qualitatively supported by Fig.~\ref{fig:visual_comparison}, while unrestricted approaches introduce observable blurring artifacts, our method produces perturbations that are virtually imperceptible to the human eye, preserving texture details and color consistency.
        
    \item We further demonstrate the practical effectiveness of these attacks in Fig.~\ref{fig:vis_results}. Despite keeping the perturbation nearly imperceptible, our method can reliably steer the model’s interpretation toward the target semantics. For instance, in Fig.~\ref{fig:vis_results}(a), although the image clearly shows ``three white wolves'', UniDiffuser is induced to describe it as a ``bathroom with tiled walls''. Even for the more robust LLaVA-NeXT (Fig.~\ref{fig:vis_results}(f)), our method drives the model to generate ``laptop computers'' from an image of a salamander. These examples show that our approach achieves precise targeted control by enforcing target-aligned semantics, causing the model’s generated descriptions to follow the target prompt rather than the original visual content.

\end{enumerate}

\begin{table*}[htbp]
\centering
\caption{Quantitative performance comparison on commercial black-box VLMs. We compare methods under {Standard} ($\epsilon=8/255$), {Unrestricted}, and {Enhanced} ($\epsilon=16/255$) settings. {Ens.} denotes the Ensemble CLIP Score.}
\label{tab:commercial_results}
\renewcommand{\arraystretch}{1.2}
\setlength{\tabcolsep}{2.5pt} 
\begin{tabular}{l|l|cc|ccc|ccc|ccc}
\toprule
\multirow{2}{*}{\textbf{Setting}} & \multirow{2}{*}{\textbf{Method}} & \multicolumn{2}{c|}{\textbf{Visual Metrics}} & \multicolumn{3}{c|}{\textbf{GPT-4o}} & \multicolumn{3}{c|}{\textbf{Gemini-2.0}} & \multicolumn{3}{c}{\textbf{Claude-3.5}} \\
 & & \textbf{SSIM} & \textbf{PSNR} & Ens. & $\text{ASR}_{\text{fool}}$ & $\text{ASR}_{\text{target}}$ & Ens. & $\text{ASR}_{\text{fool}}$ & $\text{ASR}_{\text{target}}$ & Ens. & $\text{ASR}_{\text{fool}}$ & $\text{ASR}_{\text{target}}$ \\ \midrule

% === Baseline ===
\textbf{Clean} & No Attack & 1.00 & Inf & 0.3078 & 0\% & 0\% & 0.2345 & 0\% & 0\% & 0.2558 & 0\% & 0\% \\ \midrule

% === Standard Group ===
\multirow{3}{*}{\shortstack[l]{\textbf{Standard} \\ \scriptsize 100 iter \\ \scriptsize $\epsilon=8/255$}} 
 & MF-ii~\cite{AttackVLM} & \textbf{0.90} & \textbf{34.32} & 0.3559 & 8\% & 0\% & 0.3651 & 7\% & 0\% & 0.2812 & 12\% & 0\% \\
 & M-Attack~\cite{M-Attack}& \underline{0.89} & \underline{33.82} & 0.4166 & 32\% & 4\% & 0.4096 & 25\% & 3\% & 0.2913 & 19\% & 1\% \\
 & \cellcolor{Gray}\textbf{Ours} & \cellcolor{Gray}\underline{0.89} & \cellcolor{Gray}33.71 & \cellcolor{Gray}0.4699 & \cellcolor{Gray}61\% & \cellcolor{Gray}20\% & \cellcolor{Gray}0.4466 & \cellcolor{Gray}37\% & \cellcolor{Gray}8\% & \cellcolor{Gray}0.3120 & \cellcolor{Gray}25\% & \cellcolor{Gray}3\% \\ \midrule

% === Unrestricted Group ===
\textbf{Unrestricted} 
 & AdvDiffVLM~\cite{AdvDiffVLM} & 0.69 & 23.41 & 0.5101 & 61\% & 12\% & 0.4853 & 46\% & 8\% & 0.3466 & 41\% & \underline{9\%} \\ \midrule

% === Enhanced Group ===
\multirow{3}{*}{\shortstack[l]{\textbf{Enhanced} \\ \scriptsize 300 iter \\ \scriptsize $\epsilon=16/255$}} 
 & MF-ii~\cite{AttackVLM} & {0.74} & {28.79} & 0.3759 & 14\% & 0\% & 0.3715 & 9\% & 0\% & 0.2874 & 19\% & 0\% \\
 & M-Attack~\cite{M-Attack} & {0.75} & {28.79} & \underline{0.5714} & \underline{87\%} & \underline{47\%} & \underline{0.5516} & \underline{79\%} & \underline{45\%} & \underline{0.3502} & \underline{52\%} & 7\% \\
 & \cellcolor{Gray}\textbf{Ours} & \cellcolor{Gray}{0.75} & \cellcolor{Gray}{28.73} & \cellcolor{Gray}\textbf{0.6098} & \cellcolor{Gray}\textbf{97\%} & \cellcolor{Gray}\textbf{79\%} & \cellcolor{Gray}\textbf{0.6104} & \cellcolor{Gray}\textbf{89\%} & \cellcolor{Gray}\textbf{67\%} & \cellcolor{Gray}\textbf{0.4065} & \cellcolor{Gray}\textbf{67\%} & \cellcolor{Gray}\textbf{26\%} \\ 
\bottomrule
\end{tabular}%
\end{table*}

\subsection{Evaluation on Closed-Source Commercial VLMs}
\label{sec:commercial}

In this section, we assess the practical threat of our method in real-world black-box scenarios by extending our evaluation to three leading commercial VLMs: OpenAI GPT-4o, Google Gemini-2.0, and Anthropic Claude-3.5. Due to the high inference costs and rate limits associated with commercial APIs, we conduct this evaluation on a subset of 100 randomly sampled image-text pairs. To ensure a fair and rigorous assessment against these strongly aligned models, we adopt the experimental protocols established in M-Attack~\cite{M-Attack}, utilizing an ensemble of three CLIP visual encoders as surrogates.

% === Figure: Commercial Visualizations ===
\begin{figure*}[!ht]
  \centering
  \includegraphics[width=\textwidth]{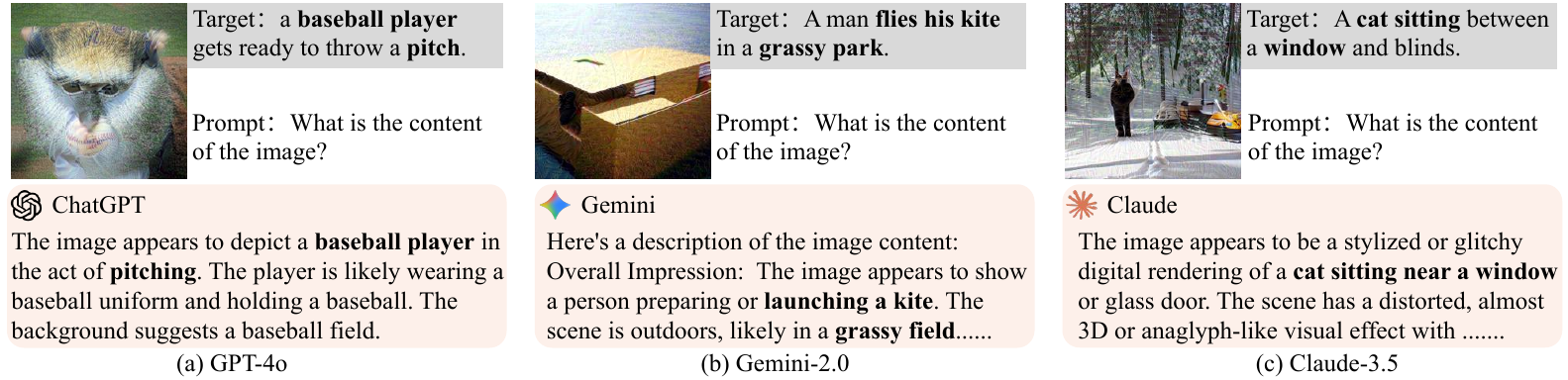}
  \caption{Visualization of targeted adversarial attacks on commercial black-box VLMs. We display the adversarial image, the target text, and the actual response generated by the model.}
  \label{fig:MLLMs_Output}
\end{figure*}

\noindent \textbf{Quantitative Attack Performance.}
We report the quantitative performance in Table \ref{tab:commercial_results}, covering both standard ($\epsilon=8/255$) and enhanced ($\epsilon=16/255$) perturbation settings to demonstrate performance gains when integrating our strategy into the M-Attack framework.

\begin{enumerate}
    \item Our method exhibits stronger effectiveness in both breaking model recognition ($\text{ASR}_{\text{fool}}$) and achieving targeted semantics ($\text{ASR}_{\text{target}}$). Under the ``Enhanced Setting'' ($\epsilon=16/255$), we achieve a 97\% $\text{ASR}_{\text{fool}}$ on GPT-4o (vs. 87\% for M-Attack), demonstrating our ability to reliably disrupt the model's visual understanding. More importantly, we translate this disruption into targeted control, boosting $\text{ASR}_{\text{target}}$ from 47\% to 79\%. This superiority extends to the robust Claude-3.5, where we surpass M-Attack in both fooling rate (67\% vs 52\%) and targeted success (26\% vs 7\%). Notably, our constrained attack even outperforms the unrestricted AdvDiffVLM (12\% success on GPT-4o) significantly, highlighting the efficiency of our semantic-guided optimization.

    \item Beyond success rates, our method consistently achieves the highest Ensemble CLIP Scores, supporting the generality of the learned hierarchical semantic alignment. For instance, under the Enhanced setting for GPT-4o, we reach an Ensemble Score of 0.6098, surpassing M-Attack (0.5714). This confirms that our hierarchical alignment ensures the target semantics are robustly recognized across diverse visual encoders, rather than overfitting to a specific surrogate.
        
\end{enumerate}

\noindent \textbf{Visual Quality Performance.}
We further provide visual evidence and imperceptibility analysis based on Figure \ref{fig:MLLMs_Output} and the visual metrics in Table \ref{tab:commercial_results}.

\begin{enumerate}
    \item As shown in the Visual Metrics column of Table \ref{tab:commercial_results}, our method maintains high structural integrity. Under the standard setting, we achieve an SSIM of 0.89 comparable to the clean baseline. Even in the enhanced setting, our method (SSIM 0.75) preserves significantly better visual quality than the unrestricted AdvDiffVLM (SSIM 0.69), which suffers from severe distortions. This quantitatively demonstrates that our method avoids the blurring artifacts typical of diffusion-based attacks while delivering stronger transferability.

    \item Figure \ref{fig:MLLMs_Output} provides qualitative evidence of our attack's precision. We successfully decouple semantic interpretation from visual content without arousing suspicion. For instance, in Figure \ref{fig:MLLMs_Output}(a), GPT-4o incorrectly describes a monkey as a ``baseball player'' preparing to pitch. Similarly, Gemini-2.0 describes a box as a person ``launching a kite'', and Claude-3.5 perceives a ``cat sitting'' near a window. These examples confirm that our method can induce targeted descriptions while preserving the visual realism of the original images.
        
\end{enumerate}

\begin{table*}[!t]
\centering
\caption{Defense-aware black-box attacks against victim VLMs. We evaluate Ens., $\text{ASR}_{\text{fool}}$, and $\text{ASR}_{\text{target}}$ under four preprocessing-based defenses: Bit Reduction, JPEG Compression, ComDefend, and DiffPure.}
\label{tab:defense_results}
\renewcommand{\arraystretch}{1.1}
% === 样式设置 ===
% ================
\resizebox{\linewidth}{!}{%
\begin{tabular}{ll|ccc|ccc|ccc|ccc}
\toprule
\multirow{2}{*}{\textbf{VLM Model}} & \multirow{2}{*}{\textbf{Attack}} & \multicolumn{3}{c|}{\textbf{Bit Reduction}\cite{Bit}} & \multicolumn{3}{c|}{\textbf{JPEG Compression}\cite{JPEG}} & \multicolumn{3}{c|}{\textbf{ComDefend}\cite{Comdefend}} & \multicolumn{3}{c}{\textbf{DiffPure}\cite{Diffpure}} \\
 &  & Ens. & $\text{ASR}_{\text{fool}}$ &  $\text{ASR}_{\text{target}}$  & Ens. & $\text{ASR}_{\text{fool}}$ &  $\text{ASR}_{\text{target}}$  & Ens. & $\text{ASR}_{\text{fool}}$ &  $\text{ASR}_{\text{target}}$  & Ens. & $\text{ASR}_{\text{fool}}$ &  $\text{ASR}_{\text{target}}$ \\ \midrule

% === UniDiffuser ===
 & AdvDiffVLM~\cite{AdvDiffVLM} & 0.5422 & 69.20\% & 21.60\% & 0.5497 & 72.20\% & 24.00\% & 0.4900 & 62.50\% & 10.50\% & 0.5110 & 62.80\% & 14.40\% \\
 & MF-ii~\cite{AttackVLM} & 0.6599 & 94.40\% & 60.00\% & 0.6316 & 91.30\% & 51.20\% & 0.5611 & 79.10\% & 28.70\% & 0.5061 & 56.00\% & 13.20\% \\
 & COA~\cite{COA} & \underline{0.6833} & \textbf{98.40\%} & \underline{65.60\%} & \underline{0.6551} & \textbf{98.40\%} & \underline{56.80\%} & \underline{0.5823} & \textbf{93.30\%} & \underline{34.50\%} & \underline{0.5447} & \textbf{83.40\%} & \underline{20.30\%} \\
 \rowcolor{Gray} \cellcolor{white} \multirow{-4}{*}{UniDiffuser} & \textbf{Ours} & \textbf{0.7452} & \underline{97.30\%} & \textbf{81.70\%} & \textbf{0.7225} & \underline{97.10\%} & \textbf{77.10\%} & \textbf{0.6505} & \underline{91.00\%} & \textbf{56.00\%} & \textbf{0.5686} & \underline{70.10\%} & \textbf{27.20\%} \\ \midrule

% === BLIP-2 ===
& AdvDiffVLM~\cite{AdvDiffVLM} & 0.5675 & 45.90\% & 19.60\% & 0.5281 & 42.50\% & 11.90\% & 0.5098 & 32.30\% & 8.10\% & \underline{0.5512} & 41.50\% & \underline{13.70\%} \\
 & MF-ii~\cite{AttackVLM} & 0.6532 & 75.50\% & 42.60\% & 0.5169 & 37.40\% & 12.40\% & 0.5095 & 31.40\% & 9.00\% & 0.5158 & 27.40\% & 7.10\% \\
 & COA~\cite{COA} & \underline{0.6912} & \underline{90.50\%} & \underline{56.40\%} & \underline{0.5446} & \textbf{60.20\%} & \underline{15.60\%} & \underline{0.5413} & \textbf{55.60\%} & \underline{14.40\%} & 0.5337 & \textbf{48.00\%} & 11.60\% \\
 \rowcolor{Gray} \cellcolor{white} \multirow{-4}{*}{BLIP-2} & \textbf{Ours} & \textbf{0.7575} & \textbf{91.90\%} & \textbf{73.90\%} & \textbf{0.5882} & \underline{56.40\%} & \textbf{27.40\%} & \textbf{0.5738} & \underline{51.80\%} & \textbf{24.30\%} & \textbf{0.5600} & \underline{43.30\%} & \textbf{18.50\%} \\ \midrule

% === InstructBLIP ===
& AdvDiffVLM~\cite{AdvDiffVLM} & 0.5645 & 48.70\% & 19.30\% & 0.5123 & 38.70\% & 11.50\% & 0.5173 & 34.30\% &  8.90\% & \underline{0.5453} & 43.00\% & \underline{15.00\%} \\
 & MF-ii~\cite{AttackVLM} & 0.6412 & 77.00\% & 43.50\% & 0.5098 & 33.70\% & 10.60\% & 0.5161 & 32.80\% & 9.70\% & 0.5126 & 27.40\% & 7.60\% \\
 & COA~\cite{COA} & \underline{0.6830} & \underline{92.20\%} & \underline{56.00\%} & \underline{0.5403} & \textbf{59.30\%} & \underline{15.70\%} & \underline{0.5420} & \textbf{55.50\%} & \underline{14.60\%} & 0.5318 & \textbf{47.80\%} & 11.60\% \\
 \rowcolor{Gray} \cellcolor{white} \multirow{-4}{*}{InstructBLIP} & \textbf{Ours} & \textbf{0.7484} & \textbf{93.30\%} & \textbf{74.40\%} & \textbf{0.5747} & \underline{54.00\%} & \textbf{25.10\%} & \textbf{0.5766} & \underline{51.70\%} & \textbf{23.30\%} & \textbf{0.5610} & \underline{44.90\%} & \textbf{18.20\%} \\ \midrule

% === MiniGPT-4 ===
& AdvDiffVLM~\cite{AdvDiffVLM} & 0.3901 & 44.70\% & 15.80\% & 0.3813 & 38.80\% & 9.80\% & 0.3833 & 31.40\% &  6.00\% & 0.3899 & 38.50\% & \underline{10.80\%} \\
 & MF-ii~\cite{AttackVLM} & 0.4547 & 76.30\% & 39.30\% & 0.3776 & 35.50\% & 8.90\% & 0.3970 & 31.10\% & 7.30\% & 0.3759 & 25.00\% & 6.40\% \\
 & COA~\cite{COA} & \underline{0.4946} & \underline{92.70\%} & \underline{52.80\%} & \underline{0.3991} & \textbf{62.60\%} & \underline{14.10\%} & \underline{0.4213} & \textbf{57.60\%} & \underline{13.00\%} & \underline{0.3910} & \textbf{48.80\%} & 9.50\% \\
 \rowcolor{Gray} \cellcolor{white} \multirow{-4}{*}{MiniGPT-4} & \textbf{Ours} & \textbf{0.5574} & \textbf{93.90\%} & \textbf{73.40\%} & \textbf{0.4325} & \underline{54.70\%} & \textbf{24.20\%} & \textbf{0.4498} & \underline{50.30\%} & \textbf{22.40\%} & \textbf{0.4202} & \underline{42.50\%} & \textbf{16.20\%} \\ \midrule

% === LLaVA ===
& AdvDiffVLM~\cite{AdvDiffVLM} & 0.4247 & 34.00\% & 6.60\% & 0.4313 & 34.40\% & 6.80\% & 0.4054 & 26.10\% &  2.10\% & 0.4248 & 31.60\% & 5.20\% \\
 & MF-ii~\cite{AttackVLM} & \underline{0.5160} & 66.20\% & \underline{30.30\%} & \underline{0.4450} & 39.40\% & \underline{11.40\%} & \underline{0.4063} & 24.50\% & \underline{4.40\%} & 0.4131 & 18.10\% & 2.60\% \\
 & COA~\cite{COA} & 0.4986 & \underline{76.70\%} & 23.20\% & 0.4137 & \underline{40.60\%} & 3.80\% & 0.4104 & \underline{35.10\%} & 2.60\% & \underline{0.4437} & \textbf{41.60\%} & \underline{10.80\%} \\
 \rowcolor{Gray} \cellcolor{white} \multirow{-4}{*}{LLaVA} & \textbf{Ours} & \textbf{0.6402} & \textbf{93.50\%} & \textbf{76.40\%} & \textbf{0.5380} & \textbf{70.10\%} & \textbf{38.10\%} & \textbf{0.4710} & \textbf{46.20\%} & \textbf{18.00\%} & \textbf{0.4520} & \underline{37.40\%} & \textbf{11.90\%} \\ \midrule

% === LLaVA-NeXT ===
& AdvDiffVLM~\cite{AdvDiffVLM} & 0.3413 & 27.40\% &  1.30\% & 0.3267 & 26.90\% & 1.90\% & \underline{0.3417} & 27.70\% & 0.40\% & 0.3294 & 24.80\% & 1.20\% \\
 & MF-ii & 0.3301 & 39.50\% & \underline{9.50\%} & \underline{0.3622} & 26.50\% & \underline{3.50\%} & 0.3276 & 20.90\% & \underline{1.30\%} & 0.3172 & 15.20\% & 0.80\% \\
 & COA~\cite{COA} & \underline{0.3701} & \underline{58.50\%} & 6.10\% & 0.3326 & \underline{32.60\%} & 1.70\% & 0.3358 & \underline{31.90\%} & 1.00\% & \underline{0.3375} & \textbf{30.20\%} & \underline{3.30\%} \\
 \rowcolor{Gray} \cellcolor{white} \multirow{-4}{*}{LLaVA-NeXT} & \textbf{Ours} & \textbf{0.4843} & \textbf{78.50\%} & \textbf{44.40\%} & \textbf{0.3954} & \textbf{49.20\%} & \textbf{15.80\%} & \textbf{0.3660} & \textbf{37.50\%} & \textbf{7.40\%} & \textbf{0.3423} & \underline{25.60\%} & \textbf{3.80\%} \\ 

\bottomrule
\end{tabular}%
}
\end{table*}

%\begin{table*}[!t]
%\centering
%\caption{Ablation study of proposed components on UniDiffuser, BLIP-2, and LLaVA. We report the Ensemble CLIP Score , $\text{ASR}_{\text{fool}}$ and $\text{ASR}_{\text{target}}$. +A, +B, and +C denote SGAI, HVSA, and CLSS, respectively. }
%\label{tab:ablation}
%% ================
%\begin{tabular}{l|ccc|ccc|ccc}
%\toprule
%\multirow{2}{*}{\textbf{Method}} & \multicolumn{3}{c|}{\textbf{UniDiffuser}} & \multicolumn{3}{c|}{\textbf{BLIP-2}} & \multicolumn{3}{c}{\textbf{LLaVA}} \\
% & Ens. & $\text{ASR}_{\text{fool}}$ &  $\text{ASR}_{\text{target}}$  & Ens. & $\text{ASR}_{\text{fool}}$ &  $\text{ASR}_{\text{target}}$ & Ens. & $\text{ASR}_{\text{fool}}$ &  $\text{ASR}_{\text{target}}$ \\ \midrule
%
%Baseline (MF-it) & 0.6650 & 91.10\% & 48.20\% & 0.7505 & 93.50\% & 70.60\% & 0.5132 & 67.70\% & 27.60\% \\
%+ A (SGAI) & 0.7545 & 99.40\% & 83.20\% & 0.7844 & 95.70\% & 83.30\% & 0.6107 & 87.70\% & 66.10\% \\
%+ B (HVSA) & 0.7393 & 99.30\% & 84.90\% & 0.8049 & 99.30\% & 92.30\% & 0.6825 & \textbf{99.90\%} & 92.90\% \\
%+ C (CLSS) & 0.7375 & 98.20\% & 78.20\% & 0.7770 & 96.30\% & 83.30\% & 0.6076 & 86.80\% & 62.60\% \\
%+ A + B & 0.7675 & 99.60\% & 88.80\% & 0.8244 & 99.30\% & 95.70\% & 0.7025 & 99.80\% & 96.20\% \\
%+ A + C & 0.7551 & 99.50\% & 83.70\% & 0.8049 & 97.10\% & 85.10\% & 0.6149 & 88.10\% & 65.30\% \\ 
%
%\rowcolor{Gray} \textbf{+ A + B + C} & \textbf{0.7705} & \textbf{99.80\%} & \textbf{89.40\%} & \textbf{0.8282} & \textbf{99.40\%} & \textbf{94.70\%} & \textbf{0.7042} & 99.60\% & \textbf{96.40\%} \\
%\bottomrule
%\end{tabular}%
%\end{table*}
%%

\begin{table*}[!t]
\centering
\caption{Ablation of the three proposed modules (SGAI, HVSA, and CLSS). We report Ens., $\text{ASR}_{\text{fool}}$, and $\text{ASR}_{\text{target}}$ on multiple VLMs. $\mathcal{L}_{txt}$ is the baseline, adding $\mathcal{L}_{anc}$, $\mathcal{L}_{feat}$, and $\mathcal{L}_{mid}$ activates SGAI, HVSA, and CLSS, respectively.}
\label{tab:ablation}
\begin{tabular}{cccc|ccc|ccc|ccc}
\toprule
\multicolumn{4}{c|}{\textbf{Loss Terms}} 
& \multicolumn{3}{c|}{\textbf{UniDiffuser}}
& \multicolumn{3}{c|}{\textbf{BLIP-2}}
& \multicolumn{3}{c}{\textbf{LLaVA}} \\
$\mathcal{L}_{txt}$ & $\mathcal{L}_{anc}$ & $\mathcal{L}_{feat}$ & $\mathcal{L}_{mid}$
& Ens. & $\text{ASR}_{\text{fool}}$ & $\text{ASR}_{\text{target}}$
& Ens. & $\text{ASR}_{\text{fool}}$ & $\text{ASR}_{\text{target}}$
& Ens. & $\text{ASR}_{\text{fool}}$ & $\text{ASR}_{\text{target}}$ \\
\midrule
\checkmark &  &  &  & 0.6650 & 91.10\% & 48.20\% & 0.7505 & 93.50\% & 70.60\% & 0.5132 & 67.70\% & 27.60\% \\
\checkmark & \checkmark &  &  & 0.7545 & 99.40\% & 83.20\% & 0.7844 & 95.70\% & 83.30\% & 0.6107 & 87.70\% & 66.10\% \\
\checkmark &  & \checkmark &  & 0.7393 & 99.30\% & 84.90\% & 0.8049 & \underline{99.30\%} & 92.30\% & 0.6825 & \textbf{99.90\%} & 92.90\% \\
\checkmark &  &  & \checkmark & 0.7375 & 98.20\% & 78.20\% & 0.7770 & 96.30\% & 83.30\% & 0.6076 & 86.80\% & 62.60\% \\
\checkmark & \checkmark & \checkmark &  & \underline{0.7675} & \underline{99.60\%} & \underline{88.80\%} & \underline{0.8244} & \underline{99.30\%} & \textbf{95.70\%} & \underline{0.7025} & \underline{99.80\%} & \underline{96.20\%} \\
\checkmark & \checkmark &  & \checkmark & 0.7551 & 99.50\% & 83.70\% & 0.8049 & 97.10\% & 85.10\% & 0.6149 & 88.10\% & 65.30\% \\ 
\rowcolor{Gray}
\checkmark & \checkmark & \checkmark & \checkmark
& \textbf{0.7705} & \textbf{99.80\%} & \textbf{89.40\%}
& \textbf{0.8282} & \textbf{99.40\%} & \underline{\textbf{94.70\%}}
& \textbf{0.7042} & 99.60\% & \textbf{96.40\%} \\
\bottomrule
\end{tabular}
\end{table*}

\subsection{Robustness Against Adversarial Defenses}

To evaluate robustness against preprocessing-based defenses, we test our method against Bit Reduction (4-bit)~\cite{Bit}, JPEG Compression (QF=75)~\cite{JPEG}, ComDefend~\cite{Comdefend}, and DiffPure~\cite{Diffpure}. As shown in Table~\ref{tab:defense_results}, our method consistently outperforms baselines in maintaining high $\text{ASR}_{\text{target}}$ and Ensemble CLIP Scores. Notably, under quantization and compression, our method significantly surpasses both MF-ii and COA across all victim models. This robustness indicates that our hierarchical alignment strategy encourages robust target-aligned representations features into the deep semantic structure of the image, allowing target semantics to survive input transformations and signal purification mechanisms that typically filter out superficial perturbations.

In comparison, while COA exhibits comparable fooling rates, its performance in targeted attacks decreases markedly under defense mechanisms. We attribute this to COA's reliance on high-frequency priors from an auxiliary network, which tends to yield high-frequency noise patterns. While effective at disrupting recognition (high $\text{ASR}_{\text{fool}}$), these patterns are less robust semantically and are easily smoothed out by purification. In contrast, our method focuses on semantic robustness through cross-modal synchronization. By constructing a robust semantic skeleton rather than relying on fragile noise, our adversarial features remain effective even after purification, yielding significantly higher targeted attack success rates compared to COA across diverse defense settings.

\subsection{Ablation of the Three Proposed Modules}

\label{sec:ablation}

We conduct ablation studies on UniDiffuser, BLIP-2, and LLaVA to examine the contribution of each component. We use MF-it as the baseline, which optimizes $\mathcal{L}_{txt}$ only. As shown in Table~\ref{tab:ablation}, enabling SGAI, HVSA, and CLSS corresponds to adding $\mathcal{L}_{anc}$, $\mathcal{L}_{feat}$, and $\mathcal{L}_{mid}$, respectively, and the full model uses all three terms together with $\mathcal{L}_{txt}$.

Overall, adding any single component on top of MF-it consistently improves targeted control and semantic consistency. In particular, introducing SGAI ($+\mathcal{L}_{anc}$) yields a large gain across all three models, e.g., on LLaVA, $\text{ASR}_{\text{target}}$ increases from 27.60\% to 66.10\%. HVSA ($+\mathcal{L}_{feat}$) provides the strongest boost on the challenging instruction-tuned model, raising $\text{ASR}_{\text{target}}$ on LLaVA to 92.90\%, and combining SGAI and HVSA further improves it to 96.20\%. CLSS ($+\mathcal{L}_{mid}$) offers complementary cross-modal guidance, improving the overall balance of Ensemble CLIP Score and targeted success when combined with the other terms. Finally, the full objective achieves the best overall performance across models, confirming that SGAI, HVSA, and CLSS are complementary for crafting transferable targeted adversarial examples.

\subsection{{Hyperparameter Selection}}
\label{sec:hyperparameter}

\begin{figure}[!t] % [t] 尝试将图片放置在页面顶部
    \centering
    \includegraphics[width=\linewidth]{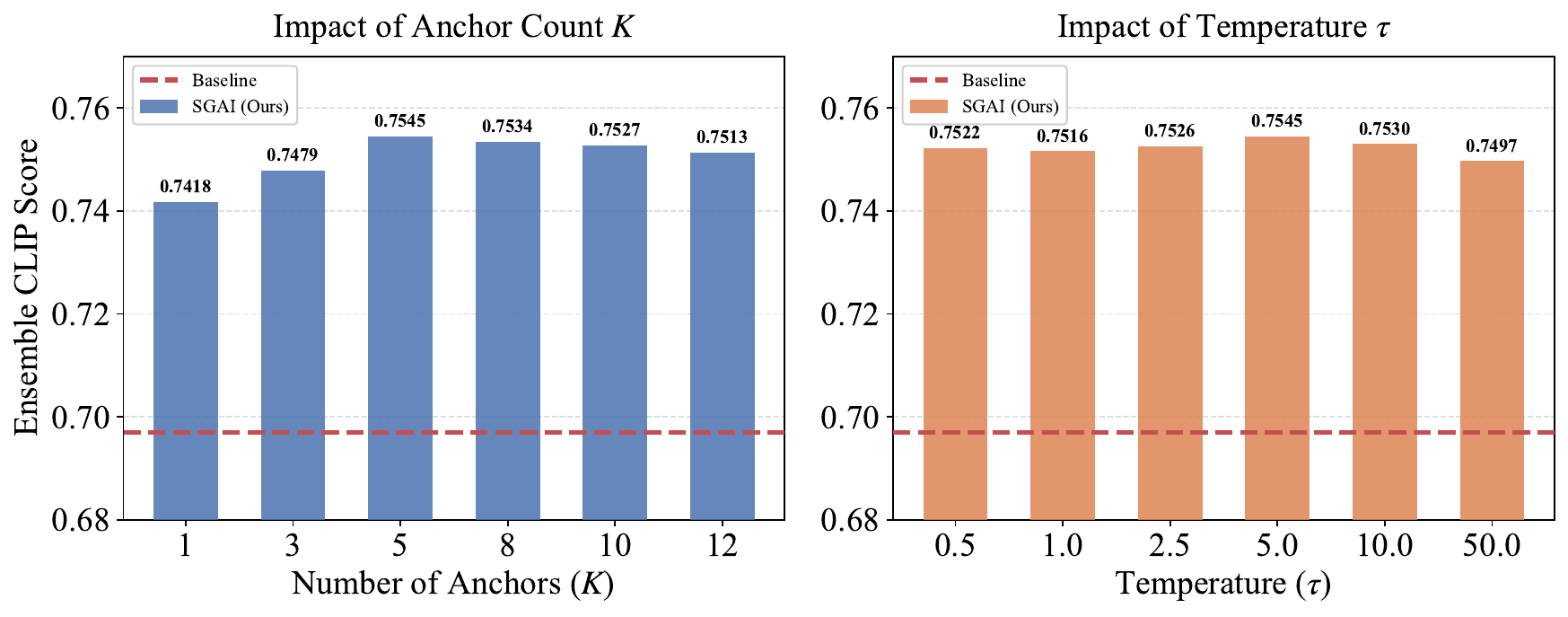} 
    \caption{Impact of hyperparameter $K$ and $\tau$ settings on attack performance evaluated on UniDiffuser.}
    \label{fig:ablation_k_tau}
\end{figure}

\noindent \textbf{Anchor Settings ($K$ and $\tau$).}
We investigate the influence of the anchor count $K$ and temperature $\tau$ on the UniDiffuser. As shown in Fig. \ref{fig:ablation_k_tau}, introducing even a single anchor ($K=1$) boosts the Ensemble CLIP Score from the baseline 0.6970 to 0.7418. The performance peaks at $K=5$ but degrades slightly with larger $K$, as excessive anchors introduce less relevant anchors. Regarding $\tau$, extreme values lead to suboptimal weighting: small $\tau$ yields an overly sharp distribution that underutilizes the semantic diversity of the anchor set, while large $\tau$ produces an overly uniform distribution, failing to emphasize the most semantically relevant anchors. The results 
find $\tau=5$ as the optimal balance. Thus, we adopt $K=5$ and $\tau=5$ as the default settings.

% ===================
\begin{figure}[!t]
    \centering
    \includegraphics[width=0.98\linewidth]{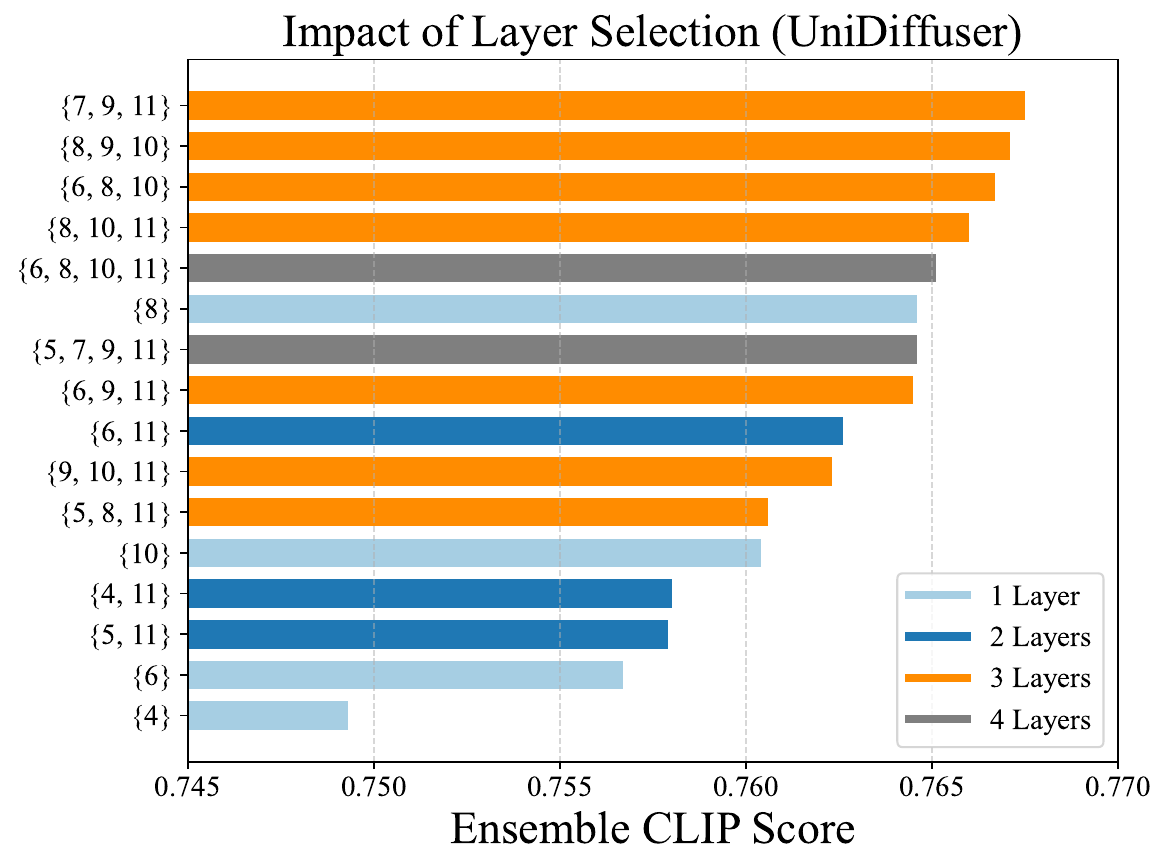}
    \caption{Ablation study on layer selection.}
    \label{fig:layer_study}
\end{figure}

\begin{figure}[!t]
    \centering
    \includegraphics[width=\linewidth]{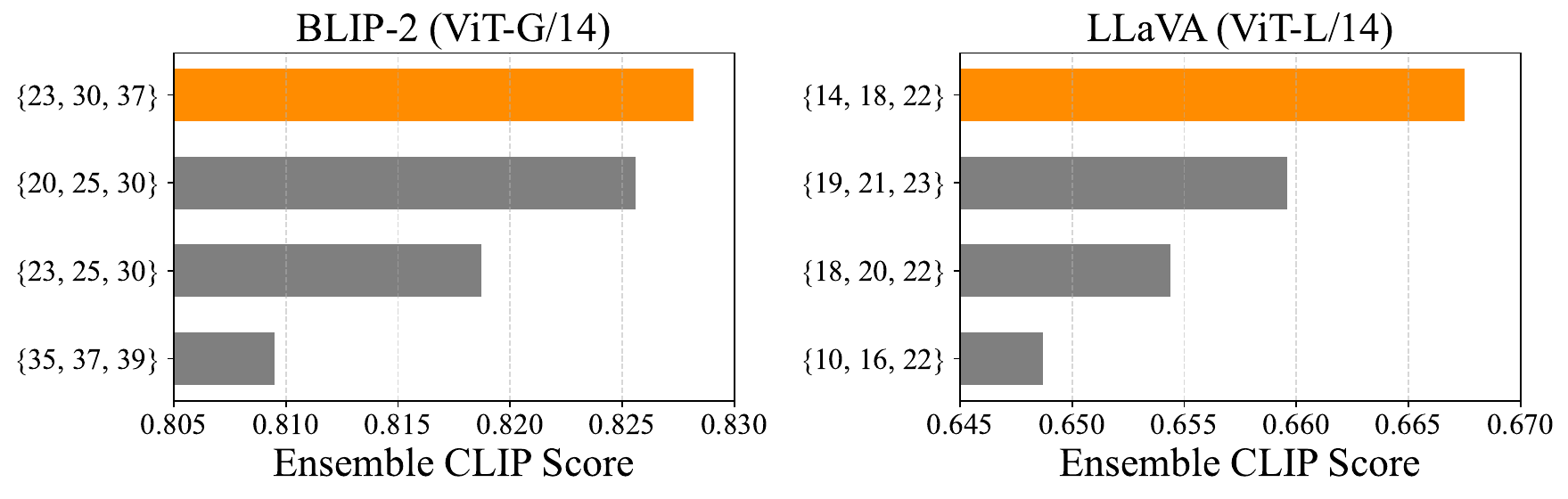}
    \caption{Validation of the strategy on different architectures.}
    \label{fig:depth_validation}
\end{figure}

%\noindent \textbf{Impact of Hierarchical Structure (Layers).} We evaluate the layer selection on UniDiffuser (ViT-B), as illustrated in Fig. \ref{fig:layer_study}. Results show that aggregating multi-scale features consistently outperforms single-layer baselines, peaking at a 3-layer combination. Furthermore, confining the alignment to a dense cluster of the deepest layers proves suboptimal, as it fails to guide the progressive semantic abstraction. Instead, a dispersed selection across the high-level semantic stages (i.e., the latter half of the network) yields the best transferability, with layers {7, 9, 11} achieving the highest score. Based on this observation, we justify our {deep-layer uniform sampling} strategy, which maps to uniform indices in the upper network blocks (specifically {7, 9, 11} for ViT-B, {14, 18, 22} for ViT-L, and {23, 30, 37} for ViT-G). This configuration ensures robust alignment across the progressive semantic abstraction process. To verify this generalization, additional experiments on BLIP-2 and LLaVA (as shown in Fig. \ref{fig:depth_validation}) confirm that these layer combinations consistently outperform other heuristic choices, demonstrating that our strategy effectively captures the optimal semantic features across diverse architectures.

\noindent \textbf{Hierarchical Structure (Layers).} We evaluate layer selection on UniDiffuser (ViT-B) in Fig.~\ref{fig:layer_study}. Multi-layer alignment consistently outperforms single-layer baselines, among the tested combinations, using 3-layers achieves the best performance. Aligning only the deepest consecutive layers is suboptimal, whereas selecting layers from the latter half of the network yields the best transferability, in particular, \{7, 9, 11\} achieves the highest score. Motivated by this, we adopt a {deep-layer uniform sampling} strategy that selects uniformly spaced layers in the upper blocks (e.g., \{7, 9, 11\} for ViT-B, \{14, 18, 22\} for ViT-L, and \{23, 30, 37\} for ViT-G). Additional experiments on BLIP-2 and LLaVA (Fig.~\ref{fig:depth_validation}) show that these selections consistently outperform other heuristic choices, validating the generality of our strategy across architectures.

\subsection{Efficiency  Analysis}
We conduct a runtime efficiency comparison to evaluate the computational cost of our method against state-of-the-art baselines using a single NVIDIA A100 GPU. As shown in Table \ref{tab:runtime}, our method achieves a favorable trade-off between attack performance and efficiency during the optimization phase. Specifically, generating an adversarial image requires only 6.58 seconds. While slightly slower than the more efficient baselines MF-it and MF-ii due to the additional computations from hierarchical alignment, our method is significantly faster than other strong baselines, being approximately 2.8$\times$ faster than COA (18.36s) and nearly 10$\times$ faster than OPS (65.33s). This indicates that our method delivers state-of-the-art transferability with a modest computational overhead.

%Regarding the preparation phase, we acknowledge that the SGAI strategy involves an initialization step to generate 20 anchor images. Following the protocol in \cite{AttackVLM}, we utilize Stable Diffusion for generation. On our hardware, generating a single anchor takes approximately 3.73s, while generating the full set of 20 anchors takes about 50.86s (batch size of 10). We argue that this one-time cost is acceptable for two main reasons. First, the anchor generation is conditioned solely on the target text and is independent of the surrogate model's gradients, granting these anchors independence from the specific surrogate model. Second, once generated, the anchors can be reused across different surrogate models or hyperparameter settings without regeneration, significantly reducing the average cost per evaluation in large-scale evaluations.

\section{Conclusion}
\label{sec:conclusion}

This work demonstrates that targeted, surrogate-driven perturbations can reliably steer the outputs of diverse VLMs, and that improving cross-modal control requires going beyond {final-layer embedding} matching. Existing targeted transfer attacks commonly narrow guidance to a {single target reference} and focus optimization on the final layer, which makes the optimization sensitive to the {surrogate-specific embedding space} and leaves {intermediate semantics underutilized} when transferring across heterogeneous VLMs.

\begin{table}[!t]
\centering
\caption{Average attack runtime (seconds) per optimization step and per adversarial image.}
\label{tab:runtime}
\renewcommand{\arraystretch}{1.1}
\footnotesize
%\resizebox{\linewidth}{!}{%
\begin{tabular}{c|l|cc}
\toprule
\textbf{Category} & \textbf{Method} & \textbf{Time/step (s)} & \textbf{Time/image (s)} \\
\midrule
Unrestricted & AdvDiffVLM~\cite{AdvDiffVLM} & -- & 60.0219 \\
\midrule
\multirow{7}{*}{\shortstack{Perturbation\\Constrained}}
 & BSR~\cite{BSR} & 0.1249 & 12.5132 \\
 & OPS~\cite{OPS} & 0.6527 & 65.3270 \\
 & M-Attack~\cite{M-Attack} & 0.0729 & 7.3433 \\
 & MF-it~\cite{AttackVLM} & \textbf{0.0249} & \textbf{2.5007} \\
 & MF-ii~\cite{AttackVLM} & \underline{0.0287} & \underline{2.8812} \\
 & COA~\cite{COA} & 0.1829 & 18.3643 \\
 % 为了不破坏第一列 multirow 的背景，只给后三列加灰色背景
 & \cellcolor{Gray}\textbf{Ours} & \cellcolor{Gray}0.0656 & \cellcolor{Gray}6.5801 \\
\bottomrule
\end{tabular}
\end{table}

To overcome these issues, we develop SGHA-Attack by combining anchor-based guidance with multi-granularity alignment. Rather than optimizing toward a single reference, the attack constructs a visually grounded {reference pool} and uses a weighted mixture to provide {stable target guidance}. {To enforce hierarchical consistency}, the attack aligns intermediate visual features at both global and spatial granularities across multiple depths, and further synchronizes visual and textual features within a shared latent subspace. This ensures that cross-modal supervision is injected early in the pipeline rather than being restricted to the {final projection}.

Evaluations on both open-source and commercial black-box VLMs confirm that this design yields stronger targeted transferability and maintains robustness under common preprocessing and purification defenses. While highly effective, the improvements can be less pronounced on advanced commercial systems that incorporate proprietary defense policies and complex transformations, motivating future work on more adaptive transfer strategies for such policy-rich black-box settings. Ultimately, these results highlight the need for defenses that explicitly safeguard internal feature hierarchies against semantic-level hijacking.

\bibliographystyle{IEEEtran}
\bibliography{VLMattack2}

\end{document}